\documentclass[letterpaper]{article} 
\usepackage[margin=1in]{geometry}
\usepackage{times}  
\usepackage{helvet}  
\usepackage{courier}  
\usepackage{url}  
\usepackage{graphicx}  
\usepackage{amsmath}
\usepackage{enumerate}
\usepackage{booktabs}
\usepackage{placeins}
\usepackage{caption}
\usepackage{subcaption}
\usepackage{xr}
\usepackage{color}
\usepackage{natbib}
\usepackage{float}
\usepackage{verbatim}
\usepackage{chngcntr}
\usepackage{hyperref}
\usepackage[title]{appendix}

\makeatletter

\let\c@table\c@figure
\makeatother 

\begin{document}
%
\title{Word Embeddings Quantify 100 Years of Gender and Ethnic Stereotypes}

\author{Nikhil Garg\\
Stanford University\\
nkgarg@stanford.edu\\
\and
Londa Schiebinger\\
Stanford University\\
schieb@stanford.edu\\
\and
Dan Jurafsky\\
Stanford University\\
jurafsky@stanford.edu\\
\and
James Zou\\
Stanford University\\
jamesz@stanford.edu\\
}
\maketitle

\begin{abstract}
Word embeddings use vectors to represent words such that the geometry between vectors captures semantic relationship between the words. In this paper, we develop a framework to demonstrate how the temporal dynamics of the embedding can be leveraged to quantify changes in stereotypes and attitudes toward women and ethnic minorities in the 20th and 21st centuries in the United States.  We integrate word embeddings trained on 100 years of text data with the U.S. Census to show that changes in the embedding track closely with demographic and occupation shifts over time. The embedding captures global social shifts -- e.g.,  the women's movement in the 1960s and Asian immigration into the U.S -- and also illuminates how specific adjectives and occupations became more closely associated with certain populations over time. Our framework for temporal analysis of word embedding opens up a powerful new intersection between machine learning and quantitative social science. 

\end{abstract}

\noindent
\section{Introduction}

The study of gender and ethnic stereotypes is an important topic across many disciplines.  Language analysis is a standard tool used to discover, understand, and demonstrate such stereotypes
\citep{hamilton_stereotypes_1986,basow_gender:_1992,wetherell_mapping_1992,holmes_handbook_2008,coates_women_2015}. Previous literature broadly establishes that language both reflects and perpetuates cultural stereotypes. However, such studies primarily leverage human surveys~\citep{ williams_sex_1977,williams_measuring_1990}, dictionary and qualitative analysis~\citep{henley_molehill_1989}, or in-depth knowledge of different languages~\citep{hellinger_gender_2001}. These methods often require time-consuming and expensive manual analysis and may not easily scale across types of stereotypes, time periods, and languages. In this paper, we propose using word embeddings, a commonly used tool in Natural Language Processing (NLP) and Machine Learning, as a new framework to measure, quantify, and compare trends in language over time. As a specific case study, we apply this tool to study the temporal dynamics of gender and ethnic stereotypes in the 20th and 21st centuries in the U.S. 

In word embedding models, each word in a given language is assigned to a high-dimensional vector such that the geometry of the vectors captures semantic relations between the words -- e.g. vectors being closer together has been shown to correspond to more similar words \citep{collobert_natural_2011}. These models are typically trained automatically on large corpora of text, such as the collections of Google News articles or all of Wikipedia, and are known to capture relationships not found through simple co-occurrence analysis. For example, the vector for \emph{France} is close to vectors for \emph{Austria} and \emph{Italy}, and the vector for \emph{XBox} is close to that of \emph{PlayStation} \citep{collobert_natural_2011}. Beyond nearby neighbors, embeddings can also capture more global relationships between words. The difference between \emph{London} and \emph{England} -- obtained by simply subtracting these two vectors -- is parallel to the vector difference between \emph{Paris} and \emph{France}. This patterns allows embeddings to capture analogy relationships, such as \emph{London} to \emph{England} is as \emph{Paris} to \emph{France}.

Recent works in machine learning demonstrate that word embeddings also capture common stereotypes, as these stereotypes are likely to be present, even if subtly, in the large corpora of training texts \citep{bolukbasi_man_2016,caliskan_semantics_2017,zhao_men_2017,van_miltenburg_stereotyping_2016}. For example, the vector for adjective \emph{honorable} would to close to the vector for \emph{man}, whereas the vector for \emph{submissive} would be closer to \emph{woman}. These stereotypes are automatically learned by the embedding algorithm, and could be problematic if the embedding is then used for sensitive applications such as search rankings, product recommendations, or translations. An important direction of research is on developing algorithms to debias the word embeddings \citep{bolukbasi_man_2016}.

In this paper, we take a new approach. We use the word embeddings as a quantitative lens through which to study historical trends -- specifically trends in the gender and ethnic stereotypes in the 20th and 21st centuries in the United States.  
We develop a systematic framework and metrics to analyze word embeddings trained over 100 years of text corpora. We show that temporal dynamics of the word embedding capture changes in gender and ethnic stereotypes over time. In particular, we  quantify how specific biases decrease over time while other stereotypes increase. Moreover, dynamics of the embedding strongly correlate with quantifiable changes in U.S. society, such as demographic and occupation shifts. For example, major transitions in the word embedding geometry reveals changes in the descriptions of genders and ethnic groups during the women's movement in the 1960-70s and Asian American population growth in the 1960s and 1980s.

We validate our findings on external metrics and show that our results are robust to the different algorithms for training the word embeddings. Our new framework reveals and quantifies how stereotypes toward women and ethnic groups have evolved in the United States. 

Our results demonstrate that word embeddings are a powerful lens through which we can systematically quantify common stereotypes and other historical trends. Embeddings thus provide an important new quantitative metric which complements existing (more qualitative) linguistic and sociological analyses of biases. In Section~\ref{sec:frameworkvalidation}, we validate that embeddings accurately capture sociological trends by comparing associations in the embeddings with census and other externally verifiable data. In Sections~\ref{sec:gender} and \ref{sec:race} we apply the framework to quantify the change in stereotypes of women and ethnic minorities. We further discuss our findings in Section~\ref{sec:discussion} and provide additional details on the method and data in Section~\ref{sec:datamethods}.

\section{Overview of the embedding framework and validations}

In this section, we briefly describe our methods and data and then validate our findings. We focus on showing that word embeddings are an effective tool to study historical biases and stereotypes by relating measurements from these embeddings to historical census data. The consistent replication of such historical data, both in magnitude and in direction of biases, validate the use of embeddings in such work. This section extends the analysis of~\cite{bolukbasi_man_2016} and~\cite{caliskan_semantics_2017} in showing that embeddings can also be used as a \textit{comparative} tool over time as a consistent metric for various biases.

\label{sec:frameworkvalidation}
\subsection{Summary of data and methods} We now briefly describe our datasets and methods, leaving details to Section~\ref{sec:datamethods} and in Appendix Section~\ref{sec:appdata}. All of our code and embeddings are available publicly\footnote{All of our own data and analysis tools are available on GitHub at \url{https://github.com/nikhgarg/EmbeddingDynamicStereotypes}. Census data is available through the Integrated Public Use Microdata Series~\citep{steven_ruggles_integrated_2015}. We link to the sources for each embedding used in Section~\ref{sec:datamethods}.}. For contemporary snapshot analysis, we use the standard 
\textit{Google News word2vec Vectors} trained on the Google News Dataset~\citep{mikolov_efficient_2013,mikolov_distributed_2013}. For historical temporal analysis, we use previously trained \textit{Google Books/COHA} embeddings, which is a set of 9 embeddings, each trained on a decade in the 1900s, using the Corpus of Historical American English and Google Books \citep{hamilton_diachronic_2016}. As additional validation, we train, using the GLoVe algorithm~\citep{pennington_glove:_2014}, embeddings from the \textit{New York Times} Annotated Corpus~\citep{sandhaus_new_2008} for every year between 1988 and 2005. We then collate several word lists to represent each gender\footnote{There is an increasingly recognized difference between sex and gender, and thus between the words \textit{male}/\textit{female }and \textit{man}/\textit{woman}, as well as non-binary categories. In this work we limit our analysis to the two major binary categories due to technical limitations, and we do use \textit{male} and \textit{female} as part of the lists of words associated with \textit{men} and \textit{women}, respectively, when measuring gender associations. Furthermore, we use results from \cite{williams_sex_1977, williams_measuring_1990} which studies stereotypes associated with \textit{sex}.} (\textit{men}, \textit{women}) and ethnicity\footnote{In this work, when we refer to \textit{Whites} or \textit{Asians}, we specifically mean the \textit{Non-Hispanic} subpopulation.} (\textit{White}, \textit{Asian}, and \textit{Hispanic}), as well as \textit{neutral} words (adjectives and occupations). For occupations, we  use historical U.S. census data~\citep{steven_ruggles_integrated_2015} to extract the proportion of workers in each occupation that belong to each gender or ethnic group and compare it to the bias in the embeddings.

Bias in the embeddings, between two groups with respect to a neutral word list, is quantified by the \textit{relative norm difference}, which is calculated as follows: (a) a representative \textit{group} vector is created as the average of the vectors for each word in the given gender/ethnicity group; (b) the average $l_2$ norm of the differences between each representative group vector and each vector in the \textit{neutral} word list of interest is calculated; (c) the relative norm difference is the difference of the average $l_2$ norms. This metric captures the relative distance (and thus relative strength of association) between the group words and the neutral word list of interest.   

\subsection{Validation of the embedding bias}

To verify that the bias in the embedding accurately reflects sociological trends, we compare the trends in the embeddings with quantifiable demographic trends in the occupation participation. We use women's participation statistics in different occupations as a benchmark because it is an objective metric of social changes.  We show that the embedding accurately captures both gender and ethic occupation proportion and consistently reflects historical changes.

\begin{figure*}[tb]
	\centering
	\begin{subfigure}{.48\linewidth}
		\centering
		\includegraphics[width=\linewidth]{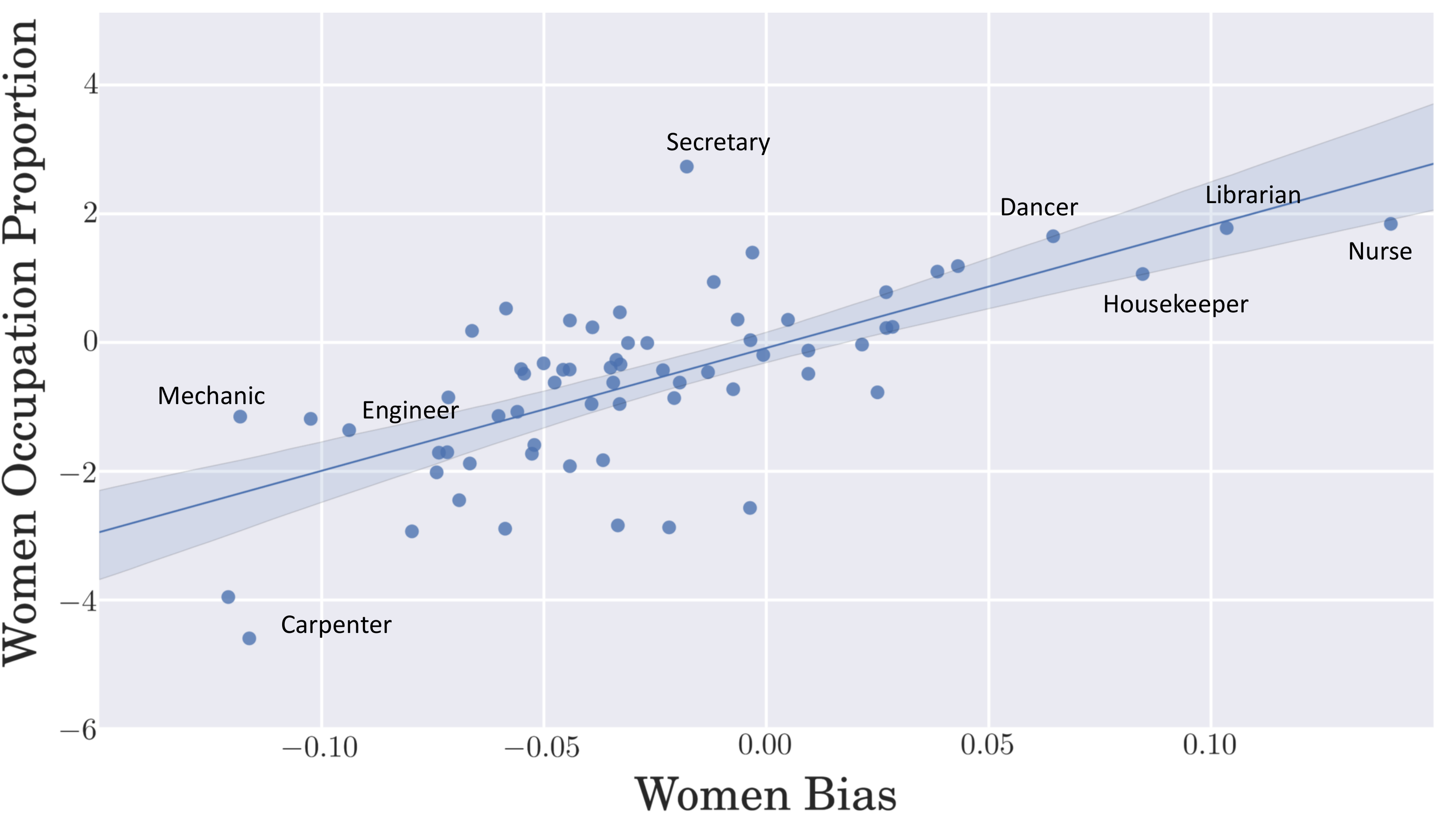}
		\caption{Woman occupation proportion vs embedding bias in Google News vectors. More positive indicates more women biased on both axes. $p<10^{-9}$, $\text{r-squared} = .462$.}
		\label{fig:occ_percents_static_scatter}
	\end{subfigure}
	\hfill
	\begin{subfigure}{.48\linewidth}
		\centering
		\includegraphics[width=\linewidth]{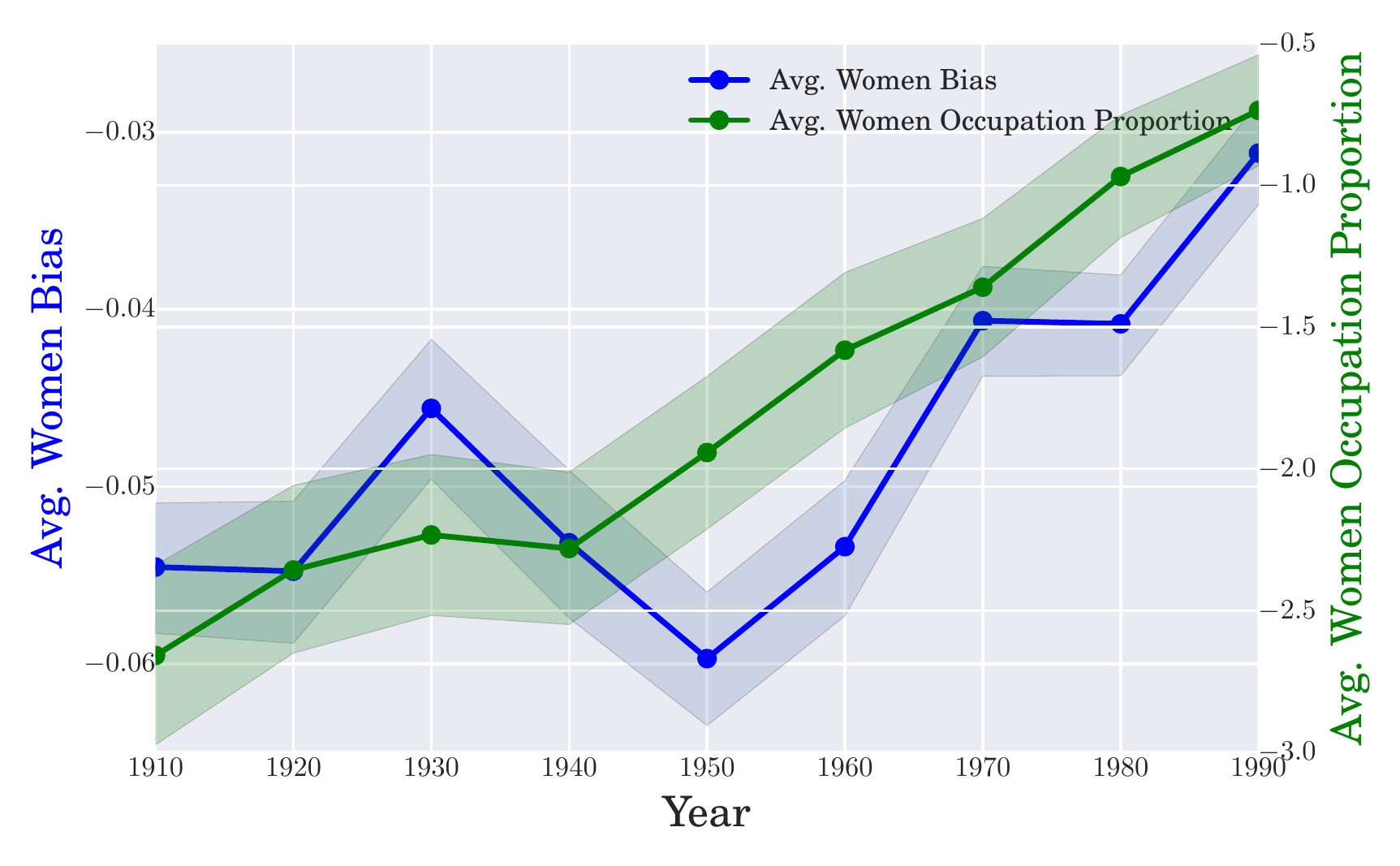}
		\caption{Average gender bias score over time in COHA embeddings in occupations vs the average log proportion. In blue is relative women bias in the embeddings, and in green is the average log proportion of women in the same occupations. }
		\label{fig:gender_bias_over_time}
	\end{subfigure}\hfill
	
		\begin{subfigure}{.48\linewidth}
					\centering
		\begin{tabular}{cc cc cc}
			Hispanic                 & Asian             &   White             \\ \hline
			housekeeper            & professor            & smith           \\
			mason            & official              & blacksmith         \\
			artist                & secretary          & surveyor          \\
			janitor             & conductor        & sheriff         \\
			dancer           & physicist             & weaver \\
			mechanic              & scientist            & administrator  \\
			photographer              & chemist           & mason   \\
			baker             & tailor              & statistician       \\
			cashier        & accountant        & clergy          \\
			driver            & engineer            & photographer        
		\end{tabular}
		\caption{The top ten occupations most closely associated with each ethnic group in the Google News embedding.}
		\label{tab:occadjstatic_threeway}
	\end{subfigure}\hfill
	\begin{subfigure}{.48\linewidth}
		\centering
		\includegraphics[width=.94\linewidth]{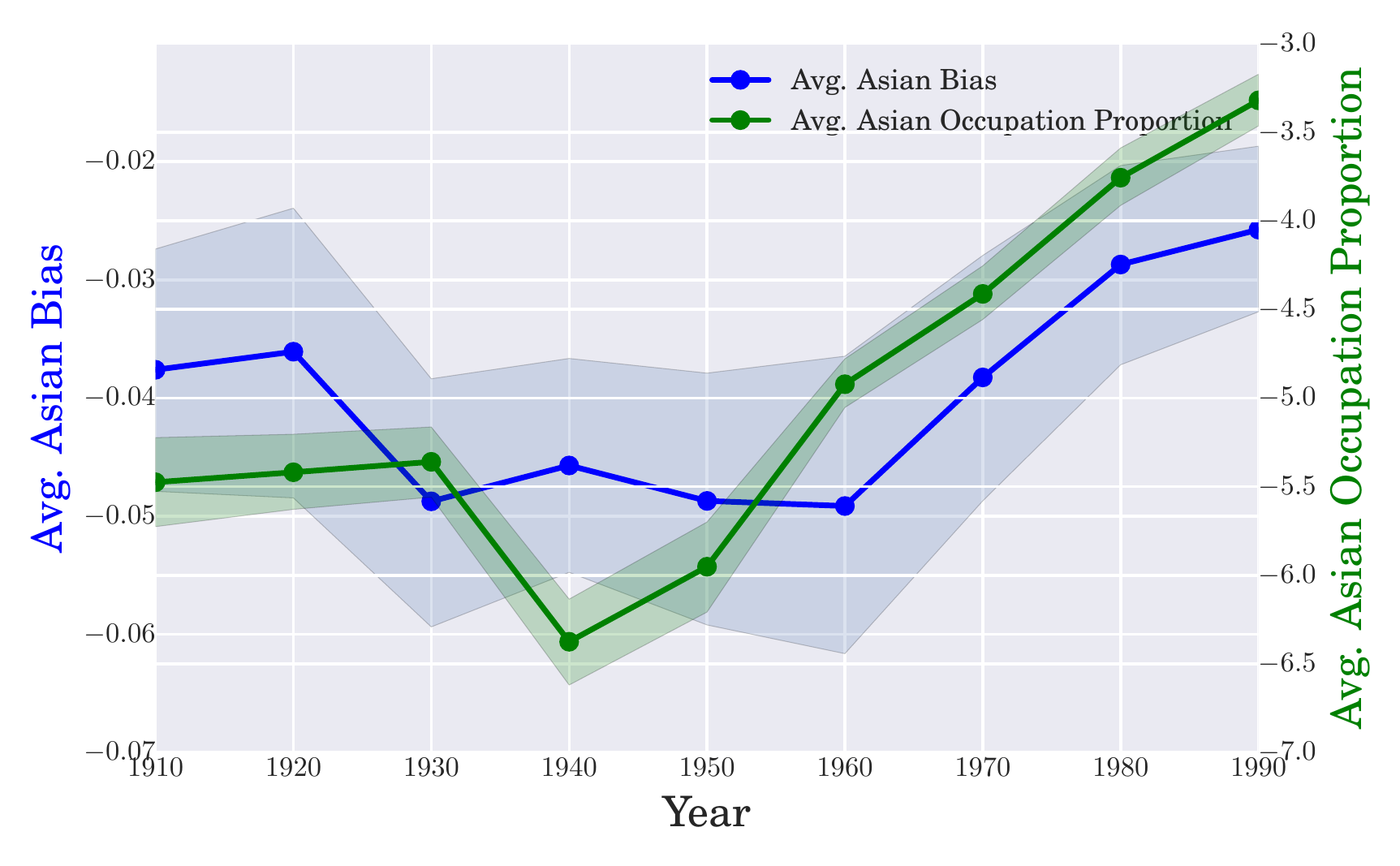}
		\caption{Average ethnic (Asian vs White) bias score over time for occupations  in COHA (blue) vs the average conditional log proportion (green). }
		\label{fig:race_bias_over_time}
	\end{subfigure}

	\caption{Associations in embeddings, both in a single dataset and across time, that track with external metrics.}
\end{figure*}

\paragraph{Comparison with women occupation participation.} We investigate how the gender bias of occupations in the word embeddings relates to the empirical proportion of women in each of these occupations in the U.S. Figure~\ref{fig:occ_percents_static_scatter} shows, for each occupation, the relationship between the log proportion (of women) in the occupation in 2015 and the relative norm distance between words associated with women and men in the Google News embeddings\footnote{Occupations whose 2015 percentage is not available, such as midwife, are omitted. We further note that the Google News embedding is trained on a corpus over time, and so the 2015 occupations are not an exact comparison.}. The relative distance in the embeddings significantly correlates with the occupation proportion ($p<10^{-9}$ with $r^2 = .46)$. It is interesting to note that the regression line goes through the origin: occupations that are close to 50-50 in its gender participation have no measurable embedding bias. This suggests that the embedding bias correctly matches the magnitude of the occupation frequency, not only which gender is more common in the occupation. 

We ask whether the relationship between embedding and occupation proportion holds true for specific occupations. We perform the same embedding bias vs. occupation frequency analysis on a subset of occupations that are deemed `professional' (e.g. \textit{nurse}, \textit{engineer}, \textit{judge}; full list in Appendix Section~\ref{sec:appdataneutralwords}), and find nearly identical correlation. We further validate this association using different embeddings trained on Wikipedia  and Common Crawl texts instead of Google News; see Appendix Section~\ref{sec:genderappstatic} for details.

Google News embedding reveals one aggregate snapshot of the bias since it is trained over a pool of news articles. We would like to also validate that for a given historical period, the embedding bias  accurately reflects occupation participation. To confirm this, we analyze the embedding of each decade of COHA from 1910 to 1990 separately.  For each decade, the embedding gender bias is significantly correlated with occupation frequency ($p<.01$), and each line approximately intercepts the origin.  Moreover these correlations are very similar over the decades, suggesting that the relationship between embedding bias score and `reality,' as measured by occupation participation, is consistent over time. This consistency makes the interpretation of embedding bias more reliable. For example, a relative woman bias of $-.05$ corresponds to $\approx12\%$ of the workforce in that occupation being woman, regardless of the embedding decade. 
See Appendix Section~\ref{sec:genderappdynamic} for details.

Next, we ask whether the changes in embeddings over decades capture changes in the women's occupation participation. 

Figure~\ref{fig:gender_bias_over_time} shows the average embedding bias over the occupations over time, overlaid with the average woman occupation log proportion over time\footnote{We only include occupations for which census data is available for every decade and which are frequent enough in all embeddings. We use the linear regression mapping inferred from all the data across decades to align the scales for the embedding bias and occupation frequency (the two $Y$ axes in the plot).}.  The average bias closely tracks with the occupation proportions over time. The average bias is negative, meaning that occupations are more closely associated with men than with women. However, we see that the bias steadily moves closer to 0 from 1950s to 1990s, suggesting that the bias is decreasing. This trend tracks with the proportional increase in the woman participation in these occupations. 

\paragraph{Comparison with ethnic occupation participation} As in the case with gender, the embeddings capture externally validated ethnic bias. Table~\ref{tab:occadjstatic_threeway} shows the ten occupations that are the most biased toward Hispanic, Asian, and White last names\footnote{We adapt the relative norm distance in Equation~\eqref{eqn:relnormdif} for three groups. For each group, we compare the its norm bias with the average bias of the other groups, i.e. $\text{bias(group 1)} = \sum_w\left[\frac{1}{2}(\|w - v_2\| + \|w - v_3\|) - \|w - v_1\|\right]$. This method can lead to the same occupation being highly ranked for multiple groups, such as happens for \textit{mason}.}. The Asian American ``model minority''~\citep{osajima_asian_2005,fong_contemporary_2002} stereotype appears predominantly; academic positions such as \textit{professor}, \textit{scientist}, and \textit{physicist} all appear as among the top Asian biased occupations. Similarly, White and Hispanic stereotypes also appear in their respective lists\footnote{\textit{Smith}, besides being an occupation, is a common last White-American last name. It is thus excluded from regressions, as are occupations such as \textit{conductor}, which have multiple meanings (train conductors as well as music conductors)}. 
As in the case with gender, the embedding bias scores are significantly correlated with the ethnic group's proportion of the occupation as measured by the U.S. Census. For Hispanics, the bias score is a significant predictor of occupation percentage at $p < 10^{-5}$; for Asians, at $p < .05$. The corresponding scatter plots of embedding bias vs. occupation proportion is in Appendix Section~\ref{sec:raceappstatic}.

Similarly, as for gender, we track the occupation bias score over time and compare it to the occupation proportions; Figure~\ref{fig:race_bias_over_time} does so for Asian Americans, relative to Whites, in the COHA embeddings. The increase in occupation proportion across all occupations is well tracked by the bias in the embeddings. 

More detail and a similar plot with Hispanic Americans is included in Appendix Section~\ref{sec:appracedynamic}.

\section{Using embeddings to quantify historical gender stereotypes}
\label{sec:gender}

Comparisons with Census data indicate that word embeddings can reliably capture stereotypes at a particular time as well as how the stereotypes evolve over time.  We apply this framework to study trends in gender bias in society, both historically and in modern times. We first show that language today, such as that in the Google News corpora, is even \textit{more} biased than occupation data alone can account for. In addition, we show that bias, as seen through adjectives associated with men and women, has decreased over time and that the women's movement in the 1960s and 1970s especially had a systemic and drastic effect in women's portrayals in literature and culture.

\subsection{Stereotype of women's occupations} 
\label{sec:objectiveoradd}
While women's occupation proportion is highly correlated with embedding gender bias, we hypothesize that the embedding could reflect additional social stereotypes beyond what can be explained by occupation participation. 
To test this hypothesis, we leverage the gender stereotype scores of occupations, as labeled by people on Amazon Mechanical Turk and provided to us by the authors of~\citep{bolukbasi_man_2016}\footnote{List of occupations available in Appendix Section~\ref{sec:appdataneutralwords}. Note that the crowdsourcing experiment collected data for a larger list of occupations; we select the occupations for which both census data and embedding orientation is also available.}. These crowdsource scores reflect aggregate human judgement as to whether an occupation is stereotypically associated with men or women\footnote{A caveat here is that the U.S. based participants on Amazon Mechanical Turk may not represent the U.S. population.}. 
Both the crowdsource scores and the occupation log proportion are significantly correlated with the embedding bias, with $r^2 = .66$ and $r^2 = .41$, respectively. 
We performed a joint regression, with the crowdsource scores and the occupation log proportions as covariates and the embedding bias as the outcome. The crowdsource scores remain significantly associated with the embedding bias while the occupation log proportions do not (at $p<10^{-5}$, versus $p>.2$ for occupation log proportion). This result indicates that the embedding bias is more closely aligned with human stereotypes than with actual occupation participation.

We also conduct two separate regressions with occupation log proportion as the independent covariate and the embedding bias and stereotype scores as two outcomes. In these regressions, a positive (negative) residual indicates that the embedding bias or stereotype score is closer to words associated with women (men) than is to be expected given the gender proportion in the occupation. We find that the residuals between the two regressions correlate significantly (Pearson coefficient $.65$, $p<10^{-5}$). This correlation suggests that the embedding bias captures the crowdsource human stereotypes beyond that can be explained by empirical differences in occupation proportions.

Where such crowdsourcing is not possible, such as in studying historical biases, word embeddings can thus further serve as an effective measurement tool. Even though the analysis in the previous section shows a strong relationship between census data and embedding bias, it is important to note that biases beyond census data also appear in the embedding.

\subsection{Quantifying changing attitudes toward women with adjective embeddings}

We now apply the insight that embeddings can be used to make comparative statements over time to study how the description of women -- through adjectives -- in literature and the broader culture has changed over time. Using word embeddings to analyze biases in adjectives could be an especially useful new approach because the literature is lacking systematic and quantitative metrics for adjective biases. We find that -- as a whole -- portrayals have changed dramatically over time, including for the better in some measurable ways. Furthermore, we find evidence for how the women's movement in the 1960s and 1970s led to systemic change in such portrayals.

\begin{figure*}[tb]
	\centering
	\begin{subtable}{.48\linewidth}
	\centering
	\begin{tabular}{ccc}
		1910        & 1950        & 1990       \\\hline
		charming    & delicate    & maternal   \\
		placid      & sweet       & morbid     \\
		delicate    & charming    & artificial \\
		passionate  & transparent & physical   \\
		sweet       & placid      & caring     \\
		dreamy      & childish    & emotional  \\
		indulgent   & soft        & protective \\
		playful     & colorless   & attractive \\
		mellow      & tasteless   & soft       \\
		sentimental & agreeable   & tidy   
	\end{tabular}
	\caption{Top adjectives associated with women in 1910, 1950, and 1990 by relative norm difference in the COHA embedding.}
	\label{tab:mostwomanadjectives191019501990}
\end{subtable}\hfill
		\begin{subfigure}{.48\linewidth}
				\centering
		\includegraphics[width=\linewidth]{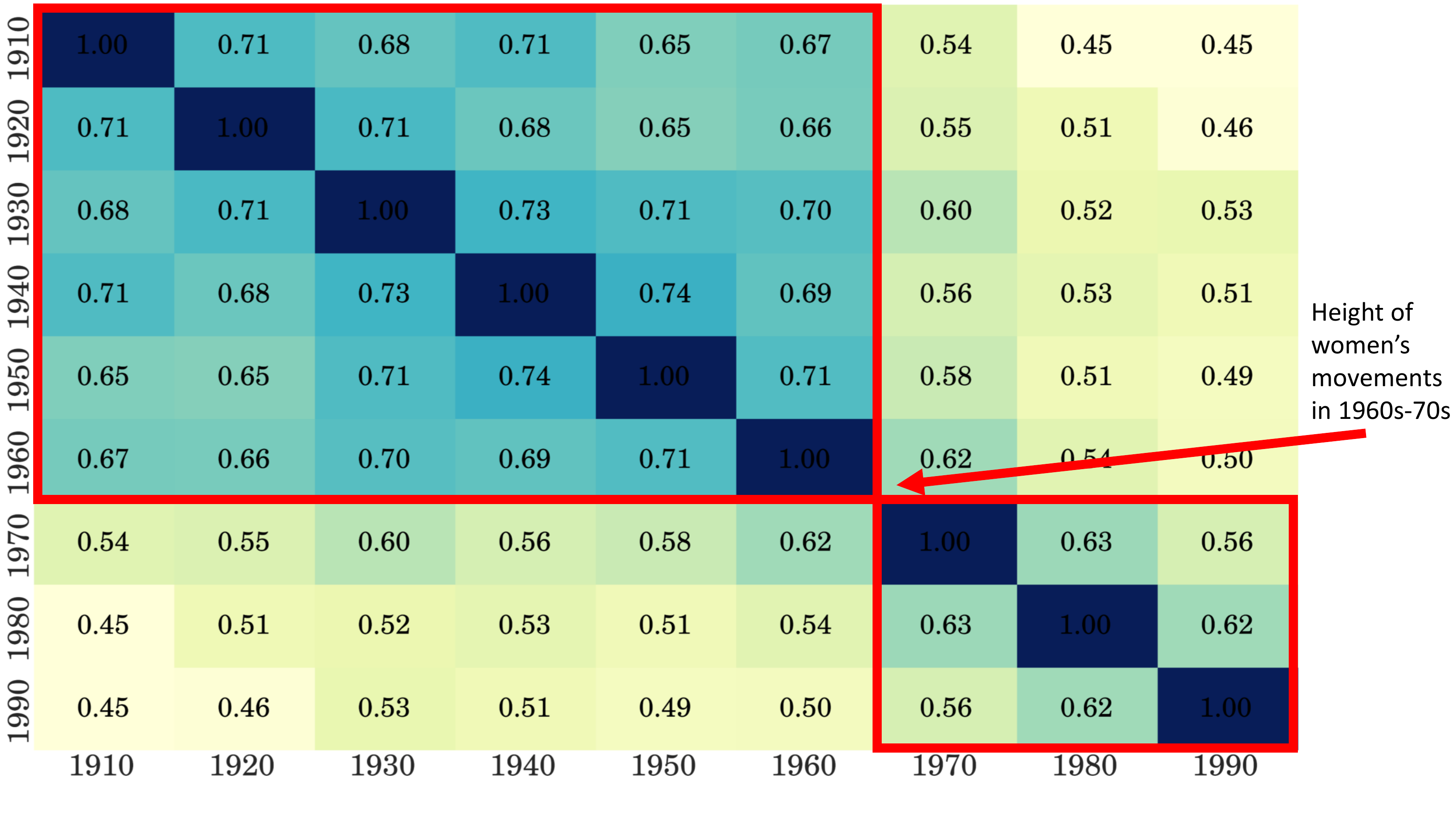}
		\caption{Pearson correlation in embedding bias scores for adjectives over time between embeddings for each decade. The phase shift in the 1960s-70s corresponds to the U.S. women's movement.} 
		\label{fig:pearcorrelation_genderapersonality_sgns}
			\end{subfigure}
			\caption{How the association of women with certain adjectives has changed over time. Though significant, measurable change has occured, the strongest associations in 1990 still indicate certain gender stereotypes, especially when compared to the top adjectives associated with men.}
\end{figure*}

\paragraph{Analysis of how adjectives change over time.} 
A difficulty in social science is the relative dearth of historical data to systematically quantify gender stereotypes, which highlights the value of our embedding framework as a quantitative tool but also makes it challenging to directly confirm our findings on adjectives. 
We leverage the best available data, which are the sex stereotype scores assigned to a set of 230 adjectives\footnote{300 adjectives are in the original studies. 70 adjectives are discarded due to low frequencies in the COHA embeddings.} by human participants \citep{williams_sex_1977,williams_measuring_1990}. This human subject study was first performed in 1977 and then repeated in 1990. We compute the correlation between the adjective embedding biases in COHA 1970s and 1990s with the respective decade human-assigned scores. In each case, the embedding bias score is significantly correlated with the human-assigned scores (at $p<.0002$). Moreover, the regression lines nearly intersect the origin, meaning that adjectives that are rated by humans to be gender neutral also tend to be unbiased in the embeddings. Appendix~\ref{sec:genderappdynamic} contains details of the analysis. These analyses suggest that the embedding gender bias effectively captures both occupation frequencies as well as human stereotypes of adjectives. 

Using this insight, we consider a subset of the adjectives describing intelligence, such as \textit{intelligent}, \textit{logical}, and \textit{thoughtful} (see Appendix~\ref{sec:appdataneutralwords} for a full list of words). This group of words on average have increased in association with women over time (from strongly biased toward men to less so), especially after the 1960s (positive trend with $p<.005$). As a comparison, we also analyze a subset of adjectives describing physical appearance -- e.g.,  \textit{attractive}, \textit{ugly}, and \textit{fashionable} -- and the bias of these words did not change significantly over time  (null hypothesis of no trend not rejected with $p >0.2$). We note that though these trends are encouraging, the top adjectives are still potentially problematic, as displayed in Table~\ref{tab:mostwomanadjectives191019501990}.

Beyond specific adjectives, we hypothesize that comparing the embedding over time could reveal more global shifts in society.  Figure~\ref{fig:pearcorrelation_genderapersonality_sgns} shows the Pearson correlation in embedding bias scores for adjectives over time between COHA embeddings for each pair of decades. As expected, the highest correlation values are near the diagonals; embeddings are most similar to those from adjacent decades. More strikingly, the matrix exhibits two clear blocks. There is a sharp divide between the 1960s and 1970s, the height of the women's movement in the United States, during which there was a large push to end legal and social barriers for women in education and the workplace~\citep{bryson_feminist_2016,rosen_world_2013}. This divide illustrates that the \textit{intelligence} vs \textit{appearance}-based adjectives example above is part of a larger language shift. We note that the effects of the women's movement, including on inclusive language, are well documented~\citep{thorne_language_1983,eckert_language_2003,rosen_world_2000,evans_tidal_2010,hellinger_gender_2001}; this work provides a new, quantitative way to measure the rate and extent of the change. A potential extension and application of this work would be to study how various narratives and descriptions of women developed and competed over time.

\paragraph{Individual words whose biases changed over time.} The embedding also reveals interesting patterns in how individual words evolve over time in their gender association. For example, the word \textit{hysterical} used to be, until the mid 1900s, a catchall term for diagnosing mental illness in women but has since become a more general word~\citep{tasca_women_2012}; such changes are clearly reflected in the embeddings, as \textit{hysterical} fell from a top 5 woman-biased word in 1920 to not in the top 100 in 1990 in the COHA embeddings\footnote{We caution that due to the noisy nature of word embeddings, dwelling on individual word rankings in isolation is potentially problematic. For example, \textit{hysterical} is more highly associated with women in the Google News vectors than \textit{emotional}. For this reason we focus on large shifts between embeddings.}. On the other hand, \textit{emotional} becomes much more strongly associated with women over time in the embeddings, reflecting its current status as a word that is largely associated with women in a pejorative sense~\citep{sanghani_feisty_2016}.

These results together demonstrate the value and potential of leveraging embeddings to study biases over time. The embeddings capture subtle individual changes in association, as well as larger historical changes. Overall, they paint a picture of a society with decreasing but still significant gender biases. 

\section{Using embeddings to quantify historical ethnic stereotypes}
\label{sec:race}

\begin{figure*}[tb]
	\begin{subfigure}{.48\linewidth}
		\centering
		\includegraphics[width=\linewidth]{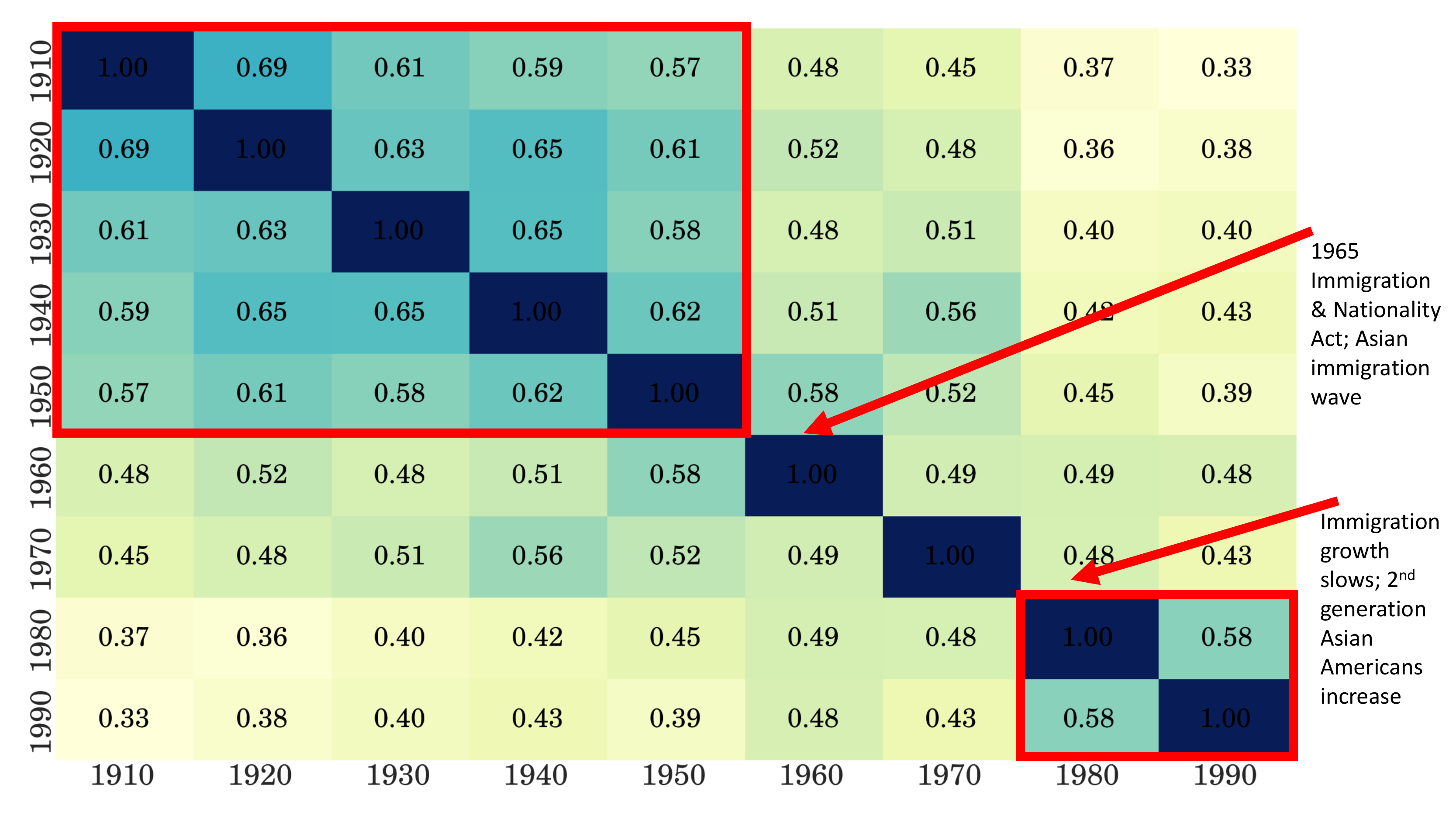}
		\caption{Pearson correlation in embedding Asian bias scores for adjectives over time between embeddings for each decade.}
		\label{fig:asianpearsoncorrelation}
	\end{subfigure}\hfill
	\begin{subtable}{.48\linewidth}
	\centering
	\begin{tabular}{ccc}
		1910       &1950   & 1990       \\\hline
irresponsible & disorganized & inhibited  \\
envious       & outrageous   & passive    \\
barbaric      & pompous      & dissolute  \\
aggressive    & unstable     & haughty    \\
transparent   & effeminate   & complacent \\
monstrous     & unprincipled & forceful   \\
hateful       & venomous     & fixed      \\
cruel         & disobedient  & active     \\
greedy        & predatory    & sensitive  \\
bizarre       & boisterous   & hearty   
	\end{tabular}
	\caption{Top Asian (vs White) Adjectives in 1910, 1950, and 1990 by relative norm difference in the COHA embedding.}
	\label{tab:mostasianadjectives19101990}
	\end{subtable}\hfill

	\begin{subfigure}{.48\linewidth}
		\centering
		\includegraphics[width=.94\linewidth]{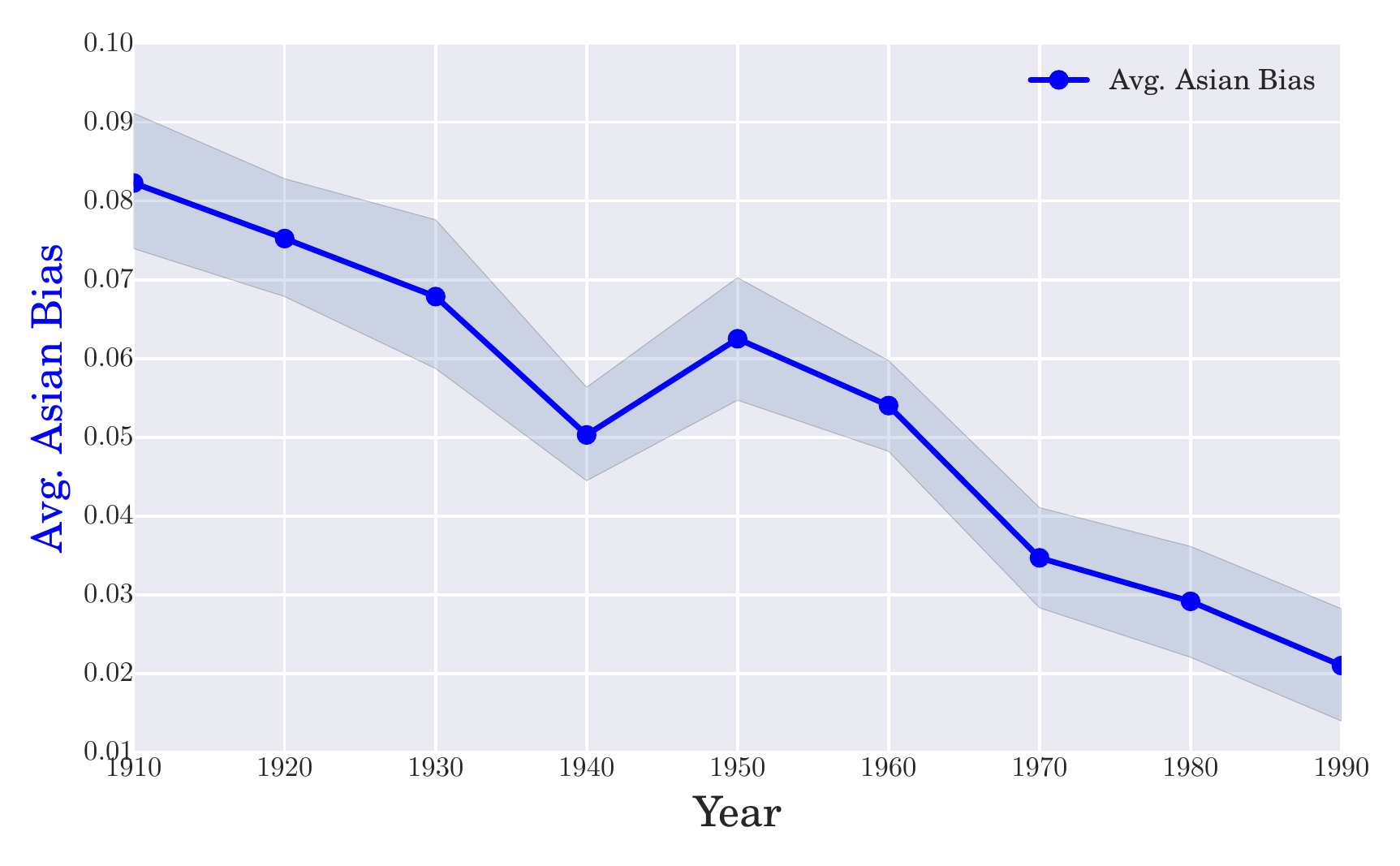}
		\caption{Asian bias score over time for words related to the outsiders in COHA data.}
		\label{fig:asiancruel_bias_over_time}
	\end{subfigure}\hfill
	\begin{subfigure}{.48\linewidth}
		\centering
		\includegraphics[width=.94\linewidth]{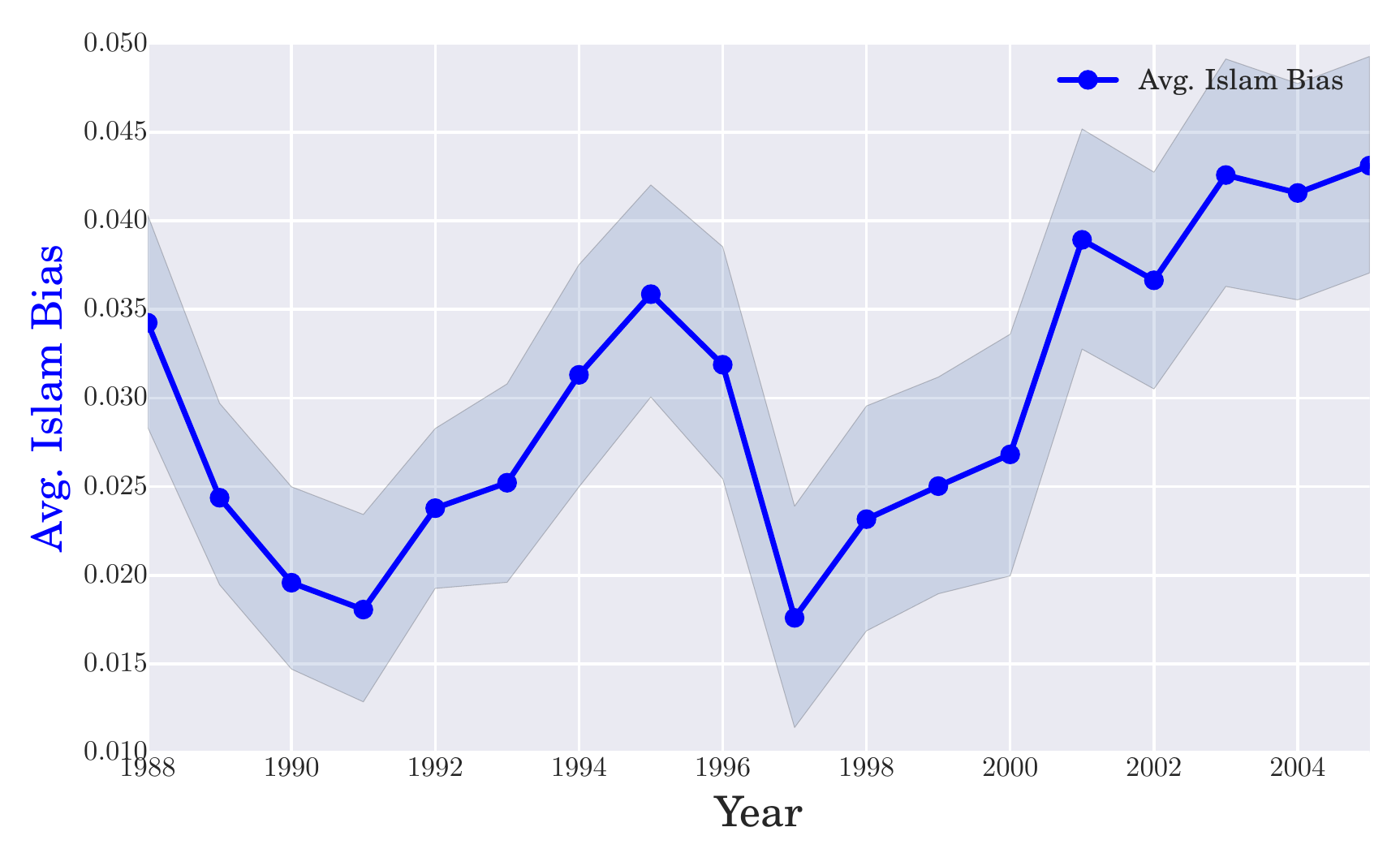}
		\caption{Religious (Islam vs Christianity) bias score over time for words related to terrorism in New York Times data. Note that embeddings are trained in 3 year windows, so, for example, \textit{2000} contains data from 1999-2001.}
		\label{fig:nytmuslim_bias_over_time}
	\end{subfigure}\hfill

	\caption{Ethnic bias in word embeddings across time.}
\end{figure*}

We now turn our attention to studying ethnic biases over time. In particular we show how immigration and broader 20th century trends broadly influenced how Asians were viewed in the U.S. We also show that embeddings can serve as effective tools to analyze finer grained trends by analyzing the portrayal of Islam in the New York Times from 1988 to 2005 in the context of terrorism. 

\subsection{Trends in Asian stereotypes}
To study Asian stereotypes in the embeddings, we use common and distinctly Asian last names, identified through a process described in Section~\ref{sec:appdatagroupwords}. This process results in a list of 20 last names that are primarily but not exclusively Chinese last names. The embeddings illustrate a dramatic story of how Asian American stereotypes developed and changed in the 20th century. Figure~\ref{fig:asianpearsoncorrelation} shows the Pearson correlation coefficient of adjective biases for each pair of embeddings over time. As with gender, the analysis shows how external events changed attitudes. There are two phase shifts in the correlation: in the 1960s, which coincides with a sharp increase in  Asian immigration into the U.S. due to the passage of the 1965 Immigration and Nationality Act, and in the 1980s, when immigration continued and the 2\textsuperscript{nd}-generation Asian-American population emerged~\citep{zong_asian_2016}.

We extract the most biased adjectives toward Asians (when compared to Whites) to gain more insights into factors driving these global changes in the embedding. Table~\ref{tab:mostasianadjectives19101990} shows the most Asian biased adjectives in 1910, 1950, and 1990. Before 1950, strongly negative words, especially those often used to describe outsiders, are among the words most associated with Asians: \textit{barbaric}, \textit{hateful}, \textit{monstrous}, \textit{bizarre}, and \textit{cruel}. However, starting around 1950 and especially by 1980, with a rising Asian population in the United States, these words are largely replaced by words often considered stereotypic of Asian-Americans today: \textit{sensitive}, \textit{passive}, \textit{complacent}, \textit{active}, and \textit{hearty}, for example~\citep{lee_behind_1994,kim_stereotypes_2002,lee_unraveling_2015}. See Table~\ref{tab:mostasianadjectives} in the Appendix for the complete list of Top 10 most Asian associated words in each decade. Using our methods regarding trends, we can quantify this change more precisely: Figure~\ref{fig:asiancruel_bias_over_time} shows the relative strength of the Asian association for words used to describe outsiders over time. As opposed to the adjectives overall, which sees 2 distinct phase shifts in Asian association, the words related to outsiders steadily decrease in Asian association over time -- including when little Chinese immigration occurred in western countries -- indicating that broader globalization trends led to changing attitudes with regards to such negative portrayals. Overall, the word embeddings exhibit a remarkable change in adjectives and attitudes toward Asian Americans during the 20th century. 

\subsection{Trends in other ethnic and cultural stereotypes}
Similar trends appear in other datasets as well. Figure~\ref{fig:nytmuslim_bias_over_time} shows, in the New York Times over two decades, how words related to Islam (vs those related to Christianity) associate with terrorism-related words. Similar to how we measure occupation related bias, we create a list of words associated with terrorism, such as \textit{terror}, \textit{bomb}, and \textit{violence}. We then measure how associated these words appear to be in the text to words representing each religion, such as \textit{mosque} and \textit{church}, for Islam and Christianity, respectively\footnote{Full word lists are available in the Appendix Section~\ref{sec:appdata}.}. Throughout the time period in the New York Times, Islam is more associated with terrorism than is Christianity. Furthermore, an increase in the association can be seen both after the 1993 World Trade Center bombings and 9/11. With a more recent dataset and using more news outlets, it would be useful to study how such attitudes have evolved since 2005.

We illustrate how word embeddings capture stereotypes toward other ethnic groups. For example, Figure~\ref{fig:russianpearsoncorrelation} in the Appendix, with Russian names, shows a dramatic shift in the 1950s, the start of the Cold War, and a minor shift during the initial years of the Russian Revolution in the 1910s-1920s. Furthermore, Figure~\ref{fig:hispanicpearsoncorrelation} in the Appendix, the correlation over time plot with Hispanic names, serves as an effective control group. It shows more steady changes in the embeddings rather than the sharp transitions found in Asian and Russian associations. This pattern is consistent with the fact that numerous events throughout the 20th century influenced the story of Hispanic immigration into the United States, with no single event playing too large a role~\citep{gutierrez_historic_2012}. 

These patterns demonstrate the usefulness of our methods to study ethnic as well as gender bias over time; similar analyses can be performed to examine shifts in the attitudes toward other ethnic groups, especially around significant global events.

\section{Discussion}
\label{sec:discussion}
In this work, we investigate how the geometry of word embeddings, with respect to gender and ethnic stereotypes, evolves over time and tracks with empirical demographic changes in the U.S.  We apply our methods to analyze word embeddings trained over 100 years of text data. In particular, we quantify the embedding biases for occupations and adjectives. Using occupations allows us to validate the method when the embedding associations are compared to empirical participation rates for each occupation. We show that both gender and ethnic occupation bias in the embeddings significantly tracks with the actual occupation frequencies. We also showed that adjective associations in the embeddings provide insight on how different groups of people are viewed over time.

As in any empirical work, the robustness of our results depends on the data sources and the metrics we choose to represent bias or association. We choose the relative norm difference metric for its simplicity, though many such metrics are reasonable. \citet{caliskan_semantics_2017}~and~\citet{bolukbasi_man_2016} leverage alternate metrics, for example. Our  metric agrees with other possible metrics -- both qualitatively through the results in the snapshot analysis for gender, which replicates prior work, and quantitatively as the metrics correlate highly one another, as shown in Appendix Section~\ref{sec:appsimmetric}. 

Another potential concern may be the dependency on our results on the specific word lists used, and that the \textit{recall} of our methods in capturing human biases may not be adequate. We take extensive care to reproduce similar results with other word lists and types of measurements to demonstrate recall. For example -- in the Appendix, we repeat the static occupation analysis using only \textit{professional} occupations and reproduce an identical Figure to~\ref{fig:occ_percents_static_scatter}. Furthermore, the plots themselves contain bootstrapped confidence intervals, i.e., the coefficients for random subsets of the occupations/adjectives, and the intervals are tight. Similarly, for adjectives, we use two different lists -- one list from~\citet{williams_sex_1977,williams_measuring_1990} for which we have labeled stereotype scores, and then a larger one for the rest of the analysis where such scores are not needed. We note that we do not tune either the embeddings or the word lists, instead opting for the largest/most general publicly available data. For reproducibility, we share our code and all word lists in a repository. That our methods replicate across many different embeddings and types of biases measured suggests its generalizability. 

A common challenge in historical analysis is that the written text in, say 1910, may not completely reflect the popular social attitude of that time. This is an important caveat to consider in interpreting the results of the embeddings trained on these earlier text corpora. The fact that the embedding bias for gender and ethnic groups does track with census proportion is a positive control that the embedding is still capturing meaningful patterns despite possible limitations in the training text. Even this control may be limited in that the census proportion does not fully capture gender or ethnic associations, even in the present day. However, the written text does serve as a window to the attitudes of the day as expressed in popular culture, and this work allows for a more systematic study of such text. 

Another limitation of our current approach is that all the embeddings used are fully ‘black-box,’ where the dimensions have no inherent meaning. In order to provide a more causal explanation of how the stereotypes appear in language, and to understand how they function, future work can leverage more recent embedding models in which certain dimensions are designed to capture various aspects of language, such as the polarity of a word or its parts of speech~\citep{rothe_word_2016}. Similarly, structural properties of words -- beyond their census information or human-rated stereotypes -- can be studied in the context of these dimensions. One can also leverage recent Bayesian embeddings models and train more fine-grained embeddings over time, rather than a separate embedding per decade as done in this work~\citep{rudolph_structured_2017,rudolph_dynamic_2017}. These approaches can be used in future work.

We view the main contribution of our work as introducing a new framework for \emph{exploring} the temporal dynamics of stereotypes through the lens of word embeddings. Our framework enables the computation of simple but quantitative measures of bias as well as easy visualizations. It is important to note that our goal is quantitative exploratory analysis rather than pinning down specific causal models of how certain stereotypes arise, though the analysis in Section~\ref{sec:objectiveoradd} makes headway on the latter and suggests that the bias found in language is \textit{more} biased than one would expect based on external, objective metrics. We believe our approach sharpens the analysis of large cultural shifts in U.S. history, e.g., the women's movement of the 1960s correlates with a sharp shift in the encoding matrix (Figure~\ref{fig:pearcorrelation_genderapersonality_sgns}) as well as changes in the biases associated with specific occupations and gender-biased adjectives (e.g., \emph{hysterical} vs \emph{emotional}). In standard quantitative social science, machine learning is used as a \emph{tool} to analyze data. Our work shows how the artifacts of machine learning (word embeddings here) can themselves be interesting objects of sociological analysis. We believe this paradigm shift can lead to many fruitful studies.

\section{Data \& Methods}
\label{sec:datamethods}
In this section we describe the datasets, embeddings and word lists used, as well as how bias is quantified. More detail, including descriptions of additional embeddings and the full word lists, are in Appendix Section~\ref{sec:appdata}. All of our data and code is available on GitHub~\footnote{\url{https://github.com/nikhgarg/EmbeddingDynamicStereotypes}}, and we link to external data sources as appropriate.	

\subsection{Embeddings}
This work uses several pre-trained word embeddings publicly available online; refer to the respective sources for in-depth discussion of their training parameters. These embeddings are among the most commonly used English embeddings, vary in the datasets on which they were trained, and between them cover the best known algorithms to construct embeddings. One  finding in this work is that, though there is some heterogeneity, gender and ethnic bias is generally consistent across embeddings. Here we restrict descriptions to embeddings used in the main exposition. For consistency, only single words are used, all vectors are normalized by their $l_2$ norm, and words are converted to lowercase. 
\paragraph{Google News word2vec Vectors} Vectors trained on about 100 billion words in the Google News Dataset \citep{mikolov_efficient_2013,mikolov_distributed_2013}. Vectors available at \url{https://code.google.com/archive/p/word2vec/}. 
\paragraph{Google Books/COHA} Trained on a combined corpus of genre-balanced Google books and the Corpus of Historical American English (COHA)~\citep{davies_400_2010} by \citeauthor{hamilton_diachronic_2016}. For each decade, a separate embedding is trained from the corpus data corresponding to that decade. The dataset is specifically designed to enable comparisons across decades, and the creators take special care to avoid selection bias issues. The vectors are available at \url{https://nlp.stanford.edu/projects/histwords/}, and we limit our analysis to the Singular Value Decomposition (SVD) and Skip-gram with Negative Sampling (SGNS, also known as word2vec) embeddings in the 1900s. Note that the Google Books data may include some non-American sources and the external metrics we use are American. However, this does not appreciably affect results. In the main text, we exclusively use SGNS embeddings; results with SVD embeddings are in the appendix and are qualitatively similar to the SGNS results. Unless otherwise specified, \textit{COHA} indicates these embeddings trained using the SGNS algorithm.
\paragraph{New York Times}
We train embeddings over time from The New York Times Annotated Corpus~\citep{sandhaus_new_2008}, using 1.8 million articles from the New York Times between 1988 and 2005. We use the GLoVe algorithm~\citep{pennington_glove:_2014}, and train embeddings over three year windows\footnote{So the \textit{2000} embeddings for example contains articles from 1999-2001.}. 

\subsection{Related Works}
Word embedding was developed as a framework to represent words as a part of the AI and natural language processing pipeline \citep{mikolov_distributed_2013}. \citep{bolukbasi_man_2016} demonstrated that word embeddings capture gender stereotypes, and \citep{caliskan_semantics_2017} additionally verified that the embedding accurately reflects human biases by comparing the embedding results with that of the Implicit Association Test. While these two papers analyzed the bias of the static Google News embedding, our paper is the first to investigate the \emph{temporal changes} in word embeddings and study how embeddings over time capture historical trends. Our paper is also the first to study attitudes toward women and ethnic minorities by quantifying the embedding of adjectives. The focus of \citep{bolukbasi_man_2016} is to develop algorithms to reduce the gender stereotype in the embedding, which is important for sensitive applications of embeddings. In contrast, our aim is not to debias, but to leverage the embedding bias to study historical changes that are otherwise challenging to quantify. \citep{caliskan_semantics_2017} shows that embeddings contain each of the associations commonly found in the Implicit Association Test. For example, European-American names are more similar to pleasant (vs unpleasant) words than are African-American names, and male names are more similar to career (vs family) words than are female names. Similarly, they show that, in the Google News embeddings, that census data corresponds to bias in the embeddings for gender.

The study of gender and ethnic stereotypes is a large focus in linguistics and sociology, and is too extensive to be surveyed here~\cite{hamilton_stereotypes_1986,basow_gender:_1992,wetherell_mapping_1992,holmes_handbook_2008,coates_women_2015}. Our main innovation is the use of word embeddings which provides a new lens to measure and quantify biases. Another related field in linguistics studies how language changes over time and has also recently employed word embeddings as a tool 
~\cite{ullmann_semantics:_1962,blank_why_1999,kulkarni_statistically_2015}. However, this literature primarily studies \textit{semantic} changes, such as how the word \textit{gay} used to primarily mean \textit{cheerful} and now means predominantly means \textit{homosexual}~\cite{hamilton_cultural_2016,hamilton_diachronic_2016}, and does not investigate bias.

\subsection{Word Lists and External Metrics}
Two types of word lists are used in this work: \textit{group} words and \textit{neutral} words. \textit{Group} words represent groups of people, such as each gender and ethnicity. \textit{Neutral} words are those that are not intrinsically gendered or ethnic\footnote{For example, \textit{fireman} or \textit{mailman} would be gendered occupation titles and so are excluded.}; relative similarities between neutral words and a pair of groups (such as men vs women) are used to measure the strength of the association in the embeddings. In this work, we use occupations and various adjective lists as neutral words. 

\paragraph{Gender} For gender, we use noun and pronoun pairs (such as \textit{he}/\textit{she}, \textit{him}/\textit{her}, etc).
\paragraph{Race/Ethnicity} To distinguish various ethnicities, we leverage the fact that the distribution of last names in the United States differs significantly by ethnicity, with the notable exception of White and Black last names. Starting with a breakdown of ethnicity by last name compiled by~\citeauthor{chalabi_dear_2014}, we identify 20 last names for each ethnicity as detailed in Appendix Section~\ref{sec:appdatagroupwords}. Our procedure, however, produces almost identical lists for White and Black Americans (with the names being mostly White by percentage), and so the analysis does not include Black Americans.  

\paragraph{Occupation Census Data}
We use occupation words for which we have gender and ethnic subgroup information over time. Group occupation percentages are obtained from the Integrated Public Use Microdata Series, part of the University of Minnesota Historical Census Project~\citep{steven_ruggles_integrated_2015}. Data coding and pre-processing are done as described in~\citep{levanon_occupational_2009}, which studies wage dynamics as women enter certain occupations over time. The IPUMS dataset includes a column, OCC1950, coding occupation census data as if it would have been coded in 1950, allowing accurate inter-year analysis. We then hand-map the occupations from this column to single-word occupations\footnote{e.g., \textit{chemical engineer} and \textit{electrical engineer} both become \textit{engineer}, and \textit{chemist} is counted as both \textit{chemist} and \textit{scientist}.}, and hand-code a subset of the occupations as \textit{professional}. In all plots containing occupation percentages for gender, the log proportion in Equation~\eqref{eqn:logprop} rather than raw percentages is used, as in~\citep{levanon_occupational_2009}. 
\begin{align}
\text{log-prop}(p) = &\log{\frac{p}{1-p}}  \label{eqn:logprop} \\\nonumber
\text{where } p= &\text{ \% of women in occupation}
\end{align}
For ethnicity, when comparing the occupation associational strength between two groups, the conditional log proportion in Equation~\eqref{eqn:logprop_conditional} is used.
\begin{align}
\text{cond-log-prop}(\text{group 1}, \text{group 2}) = \log{\frac{p}{1-p}}  \label{eqn:logprop_conditional} \\\nonumber
\text{where } p= \frac{\text{ \% of group 1}}{\text{ \% of group 1} + \text{ \% of group 2}}
\end{align}
Note that in both cases, a log proportion of $0$ indicates a balanced ratio in the occupation.

\paragraph{Occupation Gender Stereotypes}
For a limited set of occupations, we use gender stereotype scores collected from users on Amazon Mechanical Turk by \citeauthor{bolukbasi_man_2016}. These scores are compared to embedding gender association. 

\paragraph{Adjectives}
To study associations with adjectives over time, several separate lists are used. To compare adjective embedding bias to external metrics, we leverage a list of adjectives labeled by how stereotypically associated with men or women they are, as determined by a group of subjects in 1977 and 1990~\citep{williams_sex_1977,williams_measuring_1990}. For all other analyses using adjectives, a larger list of adjectives is used, primarily from~\citep{gunkel_638_1987}. Except when otherwise specified, \textit{adjectives} is used to refer to this larger list.

\subsection{Metrics}
Given two vectors, their similarity can be measured either by their negative difference norm, as in Equation~\eqref{eqn:normdif}, or by their cosine similarity, as in Equation~\eqref{eqn:cossim}. The denominators are omitted because all vectors have norm $1$.
\begin{align}
\text{neg-norm-dif}(u, v) &= -\| u-v \|_2 \label{eqn:normdif}\\
\text{cos-sim}(u, v) &= u \cdot v \label{eqn:cossim}
\end{align}

The association between the group words and neutral words is calculated as follows: construct a group vector by averaging the vectors for each word in the group; then calculate the similarity between this average vector and each word in the neutral list as above. 

The relative norm distance, which captures the relative strength of association of a set of neutral words with respect to two groups, is as described in Equation~\eqref{eqn:relnormdif}, where $M$ is the set of neutral word vectors, $v_1$ is the average vector for group one, and $v_2$ is the average vector for group two. The more positive (negative) that the relative norm distance is, the more associated the neutral words are toward group two (one). In this work, when we say that a word is \textit{biased} toward a group with respect to another group, we specifically mean in the context of the relative norm distance. \textit{Bias score} also refers to this metric.
\begin{align}
\text{relative norm distance} = \sum_{v_m \in M} \| v_m-v_1 \|_2 - \| v_m-v_2 \|_2 \label{eqn:relnormdif}
\end{align}
We can also use cosine similarity rather than the 2-norm. Appendix Section~\ref{sec:appsimmetric} shows that the choice of similarity measure is not important; the respective metrics using each similarity measure correlate highly with one another (Pearson coefficient $>.95$ in most cases). In the main text, we exclusively use the relative norm.

\FloatBarrier
\bibliographystyle{aaai}
\bibliography{bibliography}
\begin{appendices}
	
\section{Data}
This section contains information on the additional embeddings used in this supplement and the various word lists used. These lists will also be provided in a public repository alongside code to reproduce our results.

\subsection{Additional Embeddings}
\subsubsection{Wikipedia 2014+Gigaword 5 GloVe} Trained on Wikipedia data from 2014 and newswire data from the mid 1990s through 2011 \citep{parker_english_2011} using GloVe \citep{pennington_glove:_2014}. Available online at~\url{https://nlp.stanford.edu/projects/glove/}. We refer to these vectors as the Wikipedia vectors. 

\subsubsection{Common Crawl GloVe} Trained on Common Crawl using GloVe \citep{pennington_glove:_2014}. Also available online at~\url{https://nlp.stanford.edu/projects/glove/}. We refer to these vectors as the Commmon Crawl vectors. 

\label{sec:appdata}
\subsection{Group words}
\label{sec:appdatagroupwords}

\begin{description}
	\item[Man words] \textit{he, son, his, him, father, man, boy, himself, male, brother, sons, fathers, men, boys, males, brothers, uncle, uncles, nephew, nephews}
	
	\item[Woman words] \textit{she, daughter, hers, her, mother, woman, girl, herself, female, sister, daughters, mothers, women, girls, femen, sisters, aunt, aunts, niece, nieces}
	
	\item[White last names] \textit{harris, nelson, robinson, thompson, moore, wright, anderson, clark, jackson, taylor, scott, davis, allen, adams, lewis, williams, jones, wilson, martin, johnson}
	\item[Hispanic last names] \textit{ruiz, alvarez, vargas, castillo, gomez, soto, gonzalez, sanchez, rivera, mendoza, martinez, torres, rodriguez, perez, lopez, medina, diaz, garcia, castro, cruz}
	\item[Asian last names] \textit{cho, wong, tang, huang, chu, chung, ng, wu, liu, chen, lin, yang, kim, chang, shah, wang, li, khan, singh, hong}
	\item[Russian last names] \textit{gurin, minsky, sokolov, markov, maslow, novikoff, mishkin, smirnov, orloff, ivanov, sokoloff, davidoff, savin, romanoff, babinski, sorokin, levin, pavlov, rodin, agin}
	\item[Islam words] \textit{allah, ramadan, turban, emir, salaam, sunni, koran, imam, sultan, prophet, veil, ayatollah, shiite, mosque, islam, sheik, muslim, muhammad}
	\item[Christianity words] \textit{baptism, messiah, catholicism, resurrection, christianity, salvation, protestant, gospel, trinity, jesus, christ, christian, cross, catholic, church}
\end{description}

Starting with a breakdown of ethnicity by last name compiled by~\citeauthor{chalabi_dear_2014}\footnote{available \url{https://raw.githubusercontent.com/fivethirtyeight/data/master/most-common-name/surnames.csv}}, we identify 20 last names for each Whites, Asians, and Hispanics as follows: 1) Start with list of top 50 last names by percent of that ethnicity, conditioned on being top 5000 surnames overall, as well as the top 50 last names by total number in that ethnicity (i.e., multiplied count of that last name by percent in that ethnicity). 2) Choose the 20 names that appeared most on average in the Google Books/COHA vectors over time (with a minimum number for each time period). This second step ensures that an accurate ethnicity vector is identified each time period, with minimal distortions. Russian last names are collated from various sources online.

\subsection{Neutral Words}
\label{sec:appdataneutralwords}
\begin{description}
	\item[Occupations] \textit{janitor, statistician, midwife, bailiff, auctioneer, photographer, geologist, shoemaker, athlete, cashier, dancer, housekeeper, accountant, physicist, gardener, dentist, weaver, blacksmith, psychologist, supervisor, mathematician, surveyor, tailor, designer, economist, mechanic, laborer, postmaster, broker, chemist, librarian, attendant, clerical, musician, porter, scientist, carpenter, sailor, instructor, sheriff, pilot, inspector, mason, baker, administrator, architect, collector, operator, surgeon, driver, painter, conductor, nurse, cook, engineer, retired, sales, lawyer, clergy, physician, farmer, clerk, manager, guard, artist, smith, official, police, doctor, professor, student, judge, teacher, author, secretary, soldier}
	\item[Professional Occupations\footnotemark] \textit{statistician, auctioneer, photographer, geologist, accountant, physicist, dentist, psychologist, supervisor, mathematician, designer, economist, postmaster, broker, chemist, librarian, scientist, instructor, pilot, administrator, architect, surgeon, nurse, engineer, lawyer, physician, manager, official, doctor, professor, student, judge, teacher, author}
\end{description}
\footnotetext{These were hand-coded from the overall list of occupations; follow-on work should study this more systematically.}
\begin{description}
	\item[Occupations with Human Stereotype Scores from~\citep{bolukbasi_man_2016}] \textit{teacher, author, mechanic, broker, baker, surveyor, laborer, surgeon, gardener, painter, dentist, janitor, athlete, manager, conductor, carpenter, housekeeper, secretary, economist, geologist, clerk, doctor, judge, physician, lawyer, artist, instructor, dancer, photographer, inspector, musician, soldier, librarian, professor, psychologist, nurse, sailor, accountant, architect, chemist, administrator, physicist, scientist, farmer}
\end{description}
\begin{description}
	\item[Adjectives from~\citep{williams_sex_1977,williams_measuring_1990}] \textit{headstrong, thankless, tactful, distrustful, quarrelsome, effeminate, fickle, talkative, dependable, resentful, sarcastic, unassuming, changeable, resourceful, persevering, forgiving, assertive, individualistic, vindictive, sophisticated, deceitful, impulsive, sociable, methodical, idealistic, thrifty, outgoing, intolerant, autocratic, conceited, inventive, dreamy, appreciative, forgetful, forceful, submissive, pessimistic, versatile, adaptable, reflective, inhibited, outspoken, quitting, unselfish, immature, painstaking, leisurely, infantile, sly, praising, cynical, irresponsible, arrogant, obliging, unkind, wary, greedy, obnoxious, irritable, discreet, frivolous, cowardly, rebellious, adventurous, enterprising, unscrupulous, poised, moody, unfriendly, optimistic, disorderly, peaceable, considerate, humorous, worrying, preoccupied, trusting, mischievous, robust, superstitious, noisy, tolerant, realistic, masculine, witty, informal, prejudiced, reckless, jolly, courageous, meek, stubborn, aloof, sentimental, complaining, unaffected, cooperative, unstable, feminine, timid, retiring, relaxed, imaginative, shrewd, conscientious, industrious, hasty, commonplace, lazy, gloomy, thoughtful, dignified, wholesome, affectionate, aggressive, awkward, energetic, tough, shy, queer, careless, restless, cautious, polished, tense, suspicious, dissatisfied, ingenious, fearful, daring, persistent, demanding, impatient, contented, selfish, rude, spontaneous, conventional, cheerful, enthusiastic, modest, ambitious, alert, defensive, mature, coarse, charming, clever, shallow, deliberate, stern, emotional, rigid, mild, cruel, artistic, hurried, sympathetic, dull, civilized, loyal, withdrawn, confident, indifferent, conservative, foolish, moderate, handsome, helpful, gentle, dominant, hostile, generous, reliable, sincere, precise, calm, healthy, attractive, progressive, confused, rational, stable, bitter, sensitive, initiative, loud, thorough, logical, intelligent, steady, formal, complicated, cool, curious, reserved, silent, honest, quick, friendly, efficient, pleasant, severe, peculiar, quiet, weak, anxious, nervous, warm, slow, dependent, wise, organized, affected, reasonable, capable, active, independent, patient, practical, serious, understanding, cold, responsible, simple, original, strong, determined, natural, kind}
	\item[Larger adjective list, mostly from~\citep{gunkel_638_1987}] \textit{disorganized, devious, impressionable, circumspect, impassive, aimless, effeminate, unfathomable, fickle, unprincipled, inoffensive, reactive, providential, resentful, bizarre, impractical, sarcastic, misguided, imitative, pedantic, venomous, erratic, insecure, resourceful, neurotic, forgiving, profligate, whimsical, assertive, incorruptible, individualistic, faithless, disconcerting, barbaric, hypnotic, vindictive, observant, dissolute, frightening, complacent, boisterous, pretentious, disobedient, tasteless, sedentary, sophisticated, regimental, mellow, deceitful, impulsive, playful, sociable, methodical, willful, idealistic, boyish, callous, pompous, unchanging, crafty, punctual, compassionate, intolerant, challenging, scornful, possessive, conceited, imprudent, dutiful, lovable, disloyal, dreamy, appreciative, forgetful, unrestrained, forceful, submissive, predatory, fanatical, illogical, tidy, aspiring, studious, adaptable, conciliatory, artful, thoughtless, deceptive, frugal, reflective, insulting, unreliable, stoic, hysterical, rustic, inhibited, outspoken, unhealthy, ascetic, skeptical, painstaking, contemplative, leisurely, sly, mannered, outrageous, lyrical, placid, cynical, irresponsible, vulnerable, arrogant, persuasive, perverse, steadfast, crisp, envious, naive, greedy, presumptuous, obnoxious, irritable, dishonest, discreet, sporting, hateful, ungrateful, frivolous, reactionary, skillful, cowardly, sordid, adventurous, dogmatic, intuitive, bland, indulgent, discontented, dominating, articulate, fanciful, discouraging, treacherous, repressed, moody, sensual, unfriendly, optimistic, clumsy, contemptible, focused, haughty, morbid, disorderly, considerate, humorous, preoccupied, airy, impersonal, cultured, trusting, respectful, scrupulous, scholarly, superstitious, tolerant, realistic, malicious, irrational, sane, colorless, masculine, witty, inert, prejudiced, fraudulent, blunt, childish, brittle, disciplined, responsive, courageous, bewildered, courteous, stubborn, aloof, sentimental, athletic, extravagant, brutal, manly, cooperative, unstable, youthful, timid, amiable, retiring, fiery, confidential, relaxed, imaginative, mystical, shrewd, conscientious, monstrous, grim, questioning, lazy, dynamic, gloomy, troublesome, abrupt, eloquent, dignified, hearty, gallant, benevolent, maternal, paternal, patriotic, aggressive, competitive, elegant, flexible, gracious, energetic, tough, contradictory, shy, careless, cautious, polished, sage, tense, caring, suspicious, sober, neat, transparent, disturbing, passionate, obedient, crazy, restrained, fearful, daring, prudent, demanding, impatient, cerebral, calculating, amusing, honorable, casual, sharing, selfish, ruined, spontaneous, admirable, conventional, cheerful, solitary, upright, stiff, enthusiastic, petty, dirty, subjective, heroic, stupid, modest, impressive, orderly, ambitious, protective, silly, alert, destructive, exciting, crude, ridiculous, subtle, mature, creative, coarse, passive, oppressed, accessible, charming, clever, decent, miserable, superficial, shallow, stern, winning, balanced, emotional, rigid, invisible, desperate, cruel, romantic, agreeable, hurried, sympathetic, solemn, systematic, vague, peaceful, humble, dull, expedient, loyal, decisive, arbitrary, earnest, confident, conservative, foolish, moderate, helpful, delicate, gentle, dedicated, hostile, generous, reliable, dramatic, precise, calm, healthy, attractive, artificial, progressive, odd, confused, rational, brilliant, intense, genuine, mistaken, driving, stable, objective, sensitive, neutral, strict, angry, profound, smooth, ignorant, thorough, logical, intelligent, extraordinary, experimental, steady, formal, faithful, curious, reserved, honest, busy, educated, liberal, friendly, efficient, sweet, surprising, mechanical, clean, critical, criminal, soft, proud, quiet, weak, anxious, solid, complex, grand, warm, slow, false, extreme, narrow, dependent, wise, organized, pure, directed, dry, obvious, popular, capable, secure, active, independent, ordinary, fixed, practical, serious, fair, understanding, constant, cold, responsible, deep, religious, private, simple, physical, original, working, strong, modern, determined, open, political, difficult, knowledge, kind}
	\item[Intellectual Adjectives\footnotemark] \textit{precocious, resourceful, inquisitive, sagacious, inventive, astute, adaptable, reflective, discerning, intuitive, inquiring, judicious, analytical, luminous, venerable, imaginative, shrewd, thoughtful, sage, smart, ingenious, clever, brilliant, logical, intelligent, apt, genius, wise}
\end{description}
\footnotetext{mostly from \url{https://www.e-education.psu.edu/writingrecommendationlettersonline/node/151},\url{https://www.macmillandictionary.com/us/thesaurus-category/american/words-used-to-describe-intelligent-or-wise-people}}
\begin{description}
	\item [Physical Appearance Adjectives\footnotemark] \textit{alluring, voluptuous, blushing, homely, plump, sensual, gorgeous, slim, bald, athletic, fashionable, stout, ugly, muscular, slender, feeble, handsome, healthy, attractive, fat, weak, thin, pretty, beautiful, strong}
	\item [Terrorism related words] \textit{terror, terrorism, violence, attack, death, military, war, radical, injuries, bomb, target, conflict, dangerous, kill, murder, strike, dead, violence, fight, death, force, stronghold, wreckage, aggression, slaughter, execute, overthrow, casualties, massacre, retaliation, proliferation, militia, hostility, debris, acid, execution, militant, rocket, guerrilla, sacrifice, enemy, soldier, terrorist, missile, hostile, revolution, resistance, shoot}
		\item[``Other'' Adjectives] \textit{devious, bizarre, venomous, erratic, barbaric, frightening, deceitful, forceful, deceptive, envious, greedy, hateful, contemptible, brutal, monstrous, calculating, cruel, intolerant, aggressive, monstrous}
\end{description}
\footnotetext{mostly from \url{http://usefulenglish.ru/vocabulary/appearance-and-character}, \url{http://www.sightwordsgame.com/parts-of-speech/adjectives/appearance/}, \url{http://www.stgeorges.co.uk/blog/physical-appearance-adjectives-the-bald-and-the-beautiful}}

\subsection{Embedding Quality}
Here we show that the quality of the average vector constructed for each group does not appreciably change over time and thus cannot explain the overtime trends.

\begin{figure}[H]
	\centering
	\includegraphics[width=.7\linewidth]{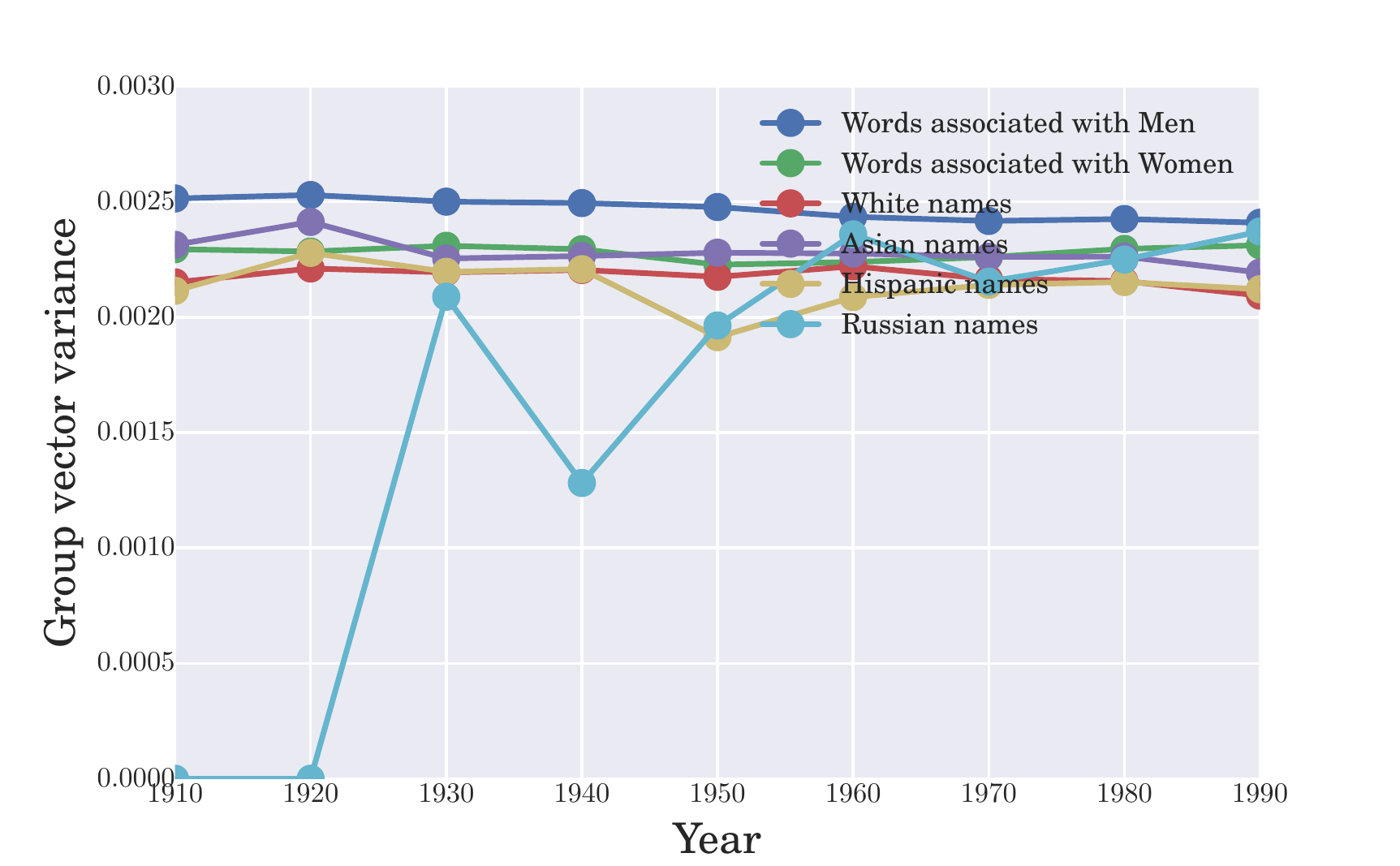}
	\caption{Average variance in each embedding dimension over time for each group in the COHA embeddings} 
	\label{fig:variancesovertime}
\end{figure}

Figure~\ref{fig:variancesovertime} shows the average frequency of words in each list across time in the COHA embeddings. Counts in general increase for all lists throughout time as datasets get larger. However, vector quality remains about the same; Figure~\ref{fig:variancesovertime} shows the average variance on each dimension across time for each group. Except for Russian names (which are used for a plot in the appendix), these variances remain relatively steady throughout time, with small potential decreases as the size of the training datasets increase, indicating that average vector quality does not change appreciably.  

\begin{figure}[H]
	\centering
	\begin{subfigure}[t]{.5\linewidth}
		\includegraphics[width=\linewidth]{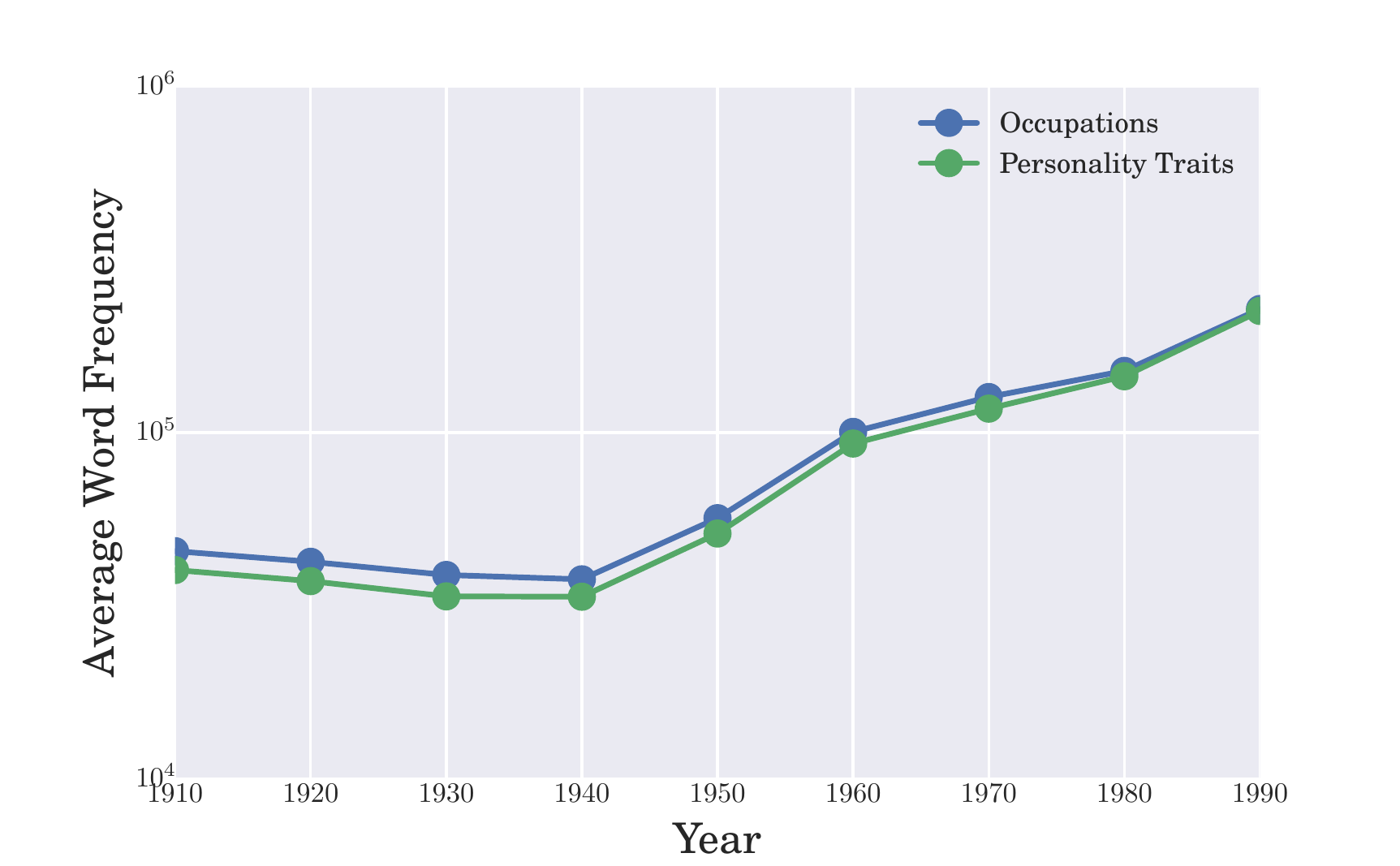}
		\caption{Neutral words} 
	\end{subfigure}\hfill
	\begin{subfigure}[t]{.5\linewidth}
		\includegraphics[width=\linewidth]{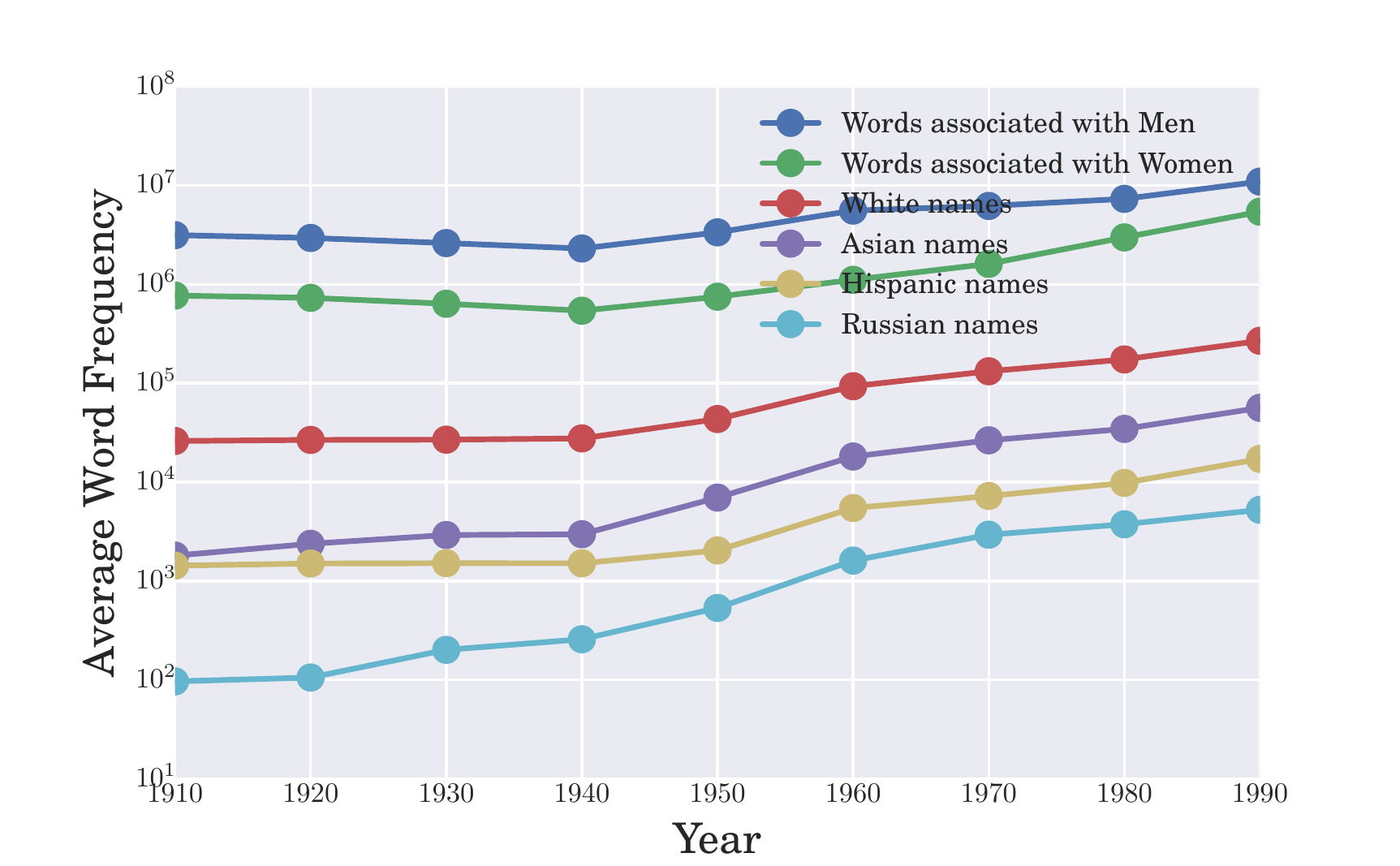}
		\caption{Group words} 
	\end{subfigure}
	\caption{Average counts of words in a list in the COHA embeddings over time}
	\label{fig:countsovertime}
\end{figure}

\subsection{Similarity metrics}
\label{sec:appsimmetric}
In the main exposition, we use the relative norm bias metric, $\sum_{v_m \in M} \| v_m-v_1 \|_2 - \| v_m-v_2 \|_2$. Using relative cosine similarity instead, $\sum_{v_m \in M} v_m \cdot v_2 - v_m \cdot v_1$,  makes no difference -- the metrics have Pearson correlation $> .95$ in general (note that Pearson correlation of $1$ is a perfectly linear relationship). The following table shows their correlation for a few embedding/neutral word/group combinations, and the pattern holds generally.

\begin{table}[H]
	\centering
	\caption{Pearson correlation of two possible bias metrics}
	\label{tab:similaritymetrics}
	\begin{tabular}{cccc}
		Embedding & Neutral Words & Groups & Pearson correlation \\ \hline
		Google News	&    Occupations        &  Men, Women     & .998\\
		Google News	&    Personality Traits        &  Men, Women    & .998 \\
		SGNS 1990	&    Occupations        & Men, Women     &.997\\
		SGNS 1990	&    Personality Traits       &  Men, Women   &.994\\
		Google News	&    Occupations        &  Whites, Asians     & .973\\
		Google News	&    Personality Traits        &  Whites, Asians    & .993 \\
		SGNS 1990	&    Occupations        & Whites, Asians     &.991\\
		SGNS 1990	&    Personality Traits       &  Whites, Asians   &.999\\
	\end{tabular}
\end{table}
We thus primarily report results using only the relative norm difference bias metric. 
\section{Gender analysis}
\subsection{Additional Validation Analysis}
\label{sec:genderappstatic}
We validate that the COHA embeddings over time are consistent with the change in bias over time. First, we show that the relationship between the embedding bias score and woman occupation proportions is approximately consistent over time, i.e. that a given embedding bias score corresponds to the \textit{same} occupation proportion regardless of decade. For example, a relative woman bias of $-.05$ corresponds to $\approx12\%$ of the workforce in that occupation being woman, regardless of the embedding decade. Given that the occupation proportions themselves shift considerably over time as women enter the work force, the mapping from embedding bias to occupation proportion is remarkably consistent over time. When a single model is trained across all years with COHA embeddings, the embedding bias score alone (without year information) is a significant predictor of occupation proportion at $p < 10^{-37}$.  

This single model performs similarly to individual models trained for each decade's embedding. Table~\ref{tab:overtimeMSE} in the Appendix, for example, illustrates that the Mean Squared Error for each decade in the aggregate model is within about $10\%$ of that in the individual models. The r-squared values of the embedding bias vs. occupation correlation in each decade are also similar to that of the aggregate correlation. Furthermore, the trained models are themselves similar: as Appendix Table~\ref{tab:overtimecoefficients} shows, there is considerable overlap in the coefficient confidence intervals for embedding bias in each individual model over time. 

Table~\ref{tab:occadjstatic} shows the top occupations and adjectives by gender in the Google News embedding.

\begin{table}[t]
	\centering
	\begin{tabular}{cc|cc}
		\multicolumn{2}{c|}{\textbf{Occupations}} & \multicolumn{2}{c}{\textbf{Adjectives}} \\
		Man                 & Woman             & Man               & Woman             \\ \hline
		carpenter            & nurse            & honorable            & maternal         \\
		mechanic            & midwife              & ascetic            & romantic               \\
		mason                & librarian          & amiable          & submissive            \\
		blacksmith             & housekeeper        & dissolute             & hysterical           \\
		retired           & dancer             & arrogant        & elegant           \\
		architect              & teacher            & erratic           & caring         \\
		engineer              & cashier          & heroic            & delicate               \\
		mathematician             & student              & boyish          & superficial    \\
		shoemaker        & designer        & fanatical             & neurotic      \\
		physicist            & weaver            & aimless             & attractive
	\end{tabular}
	\caption{Top occupations and adjectives by gender in the Google News embedding.}
	\label{tab:occadjstatic}
\end{table}

Below, we first show the regression table corresponding to Figure~\ref{fig:occ_percents_static_scatter}, and then show the same plot for other embeddings, as well as the subset of occupations corresponding to ``professional'' occupations.

\begin{table}[H]
	\begin{center}
		\begin{tabular}{lclc}
			\toprule
			\textbf{Dep. Variable:}             & Woman Occupation Proportion & \textbf{  R-squared:         } &     0.462   \\
			\textbf{Model:}                     &             OLS              & \textbf{  Adj. R-squared:    } &     0.453   \\
			\textbf{Method:}                    &        Least Squares         & \textbf{  F-statistic:       } &     54.85   \\
			\textbf{Date:}                      &       Fri, 08 Sep 2017       & \textbf{  Prob (F-statistic):} &  3.59e-10   \\
			\textbf{Time:}                      &           07:50:06           & \textbf{  Log-Likelihood:    } &   -94.098   \\
			\textbf{No. Observations:}          &                66            & \textbf{  AIC:               } &     192.2   \\
			\textbf{Df Residuals:}              &                64            & \textbf{  BIC:               } &     196.6   \\
			\textbf{Df Model:}                  &                 1            & \textbf{                     } &             \\
			\bottomrule
		\end{tabular}
		\begin{tabular}{lccccc}
			& \textbf{coef} & \textbf{std err} & \textbf{t} & \textbf{P$>$$|$t$|$} & \textbf{[95.0\% Conf. Int.]}  \\
			\midrule
			\textbf{Relative Woman Similarity} &      19.0754  &        2.576     &     7.406  &         0.000        &        13.930    24.221       \\
			\textbf{const}                      &      -0.0853  &        0.144     &    -0.593  &         0.555        &        -0.373     0.202       \\
			\bottomrule
		\end{tabular}
		\begin{tabular}{lclc}
			\textbf{Omnibus:}       &  3.174 & \textbf{  Durbin-Watson:     } &    2.346  \\
			\textbf{Prob(Omnibus):} &  0.205 & \textbf{  Jarque-Bera (JB):  } &    2.821  \\
			\textbf{Skew:}          & -0.147 & \textbf{  Prob(JB):          } &    0.244  \\
			\textbf{Kurtosis:}      &  3.969 & \textbf{  Cond. No.          } &     20.5  \\
			\bottomrule
		\end{tabular}
		\caption{OLS Regression Results corresponding to Figure~\ref{fig:occ_percents_static_scatter}, regressing woman occupation log proportion with word vector relative distance in Google News vectors in 2015.}
	\end{center}
\end{table}

\begin{figure}[H]
	\centering
	\begin{subfigure}[H]{.45\linewidth}
		\includegraphics[width=\linewidth]{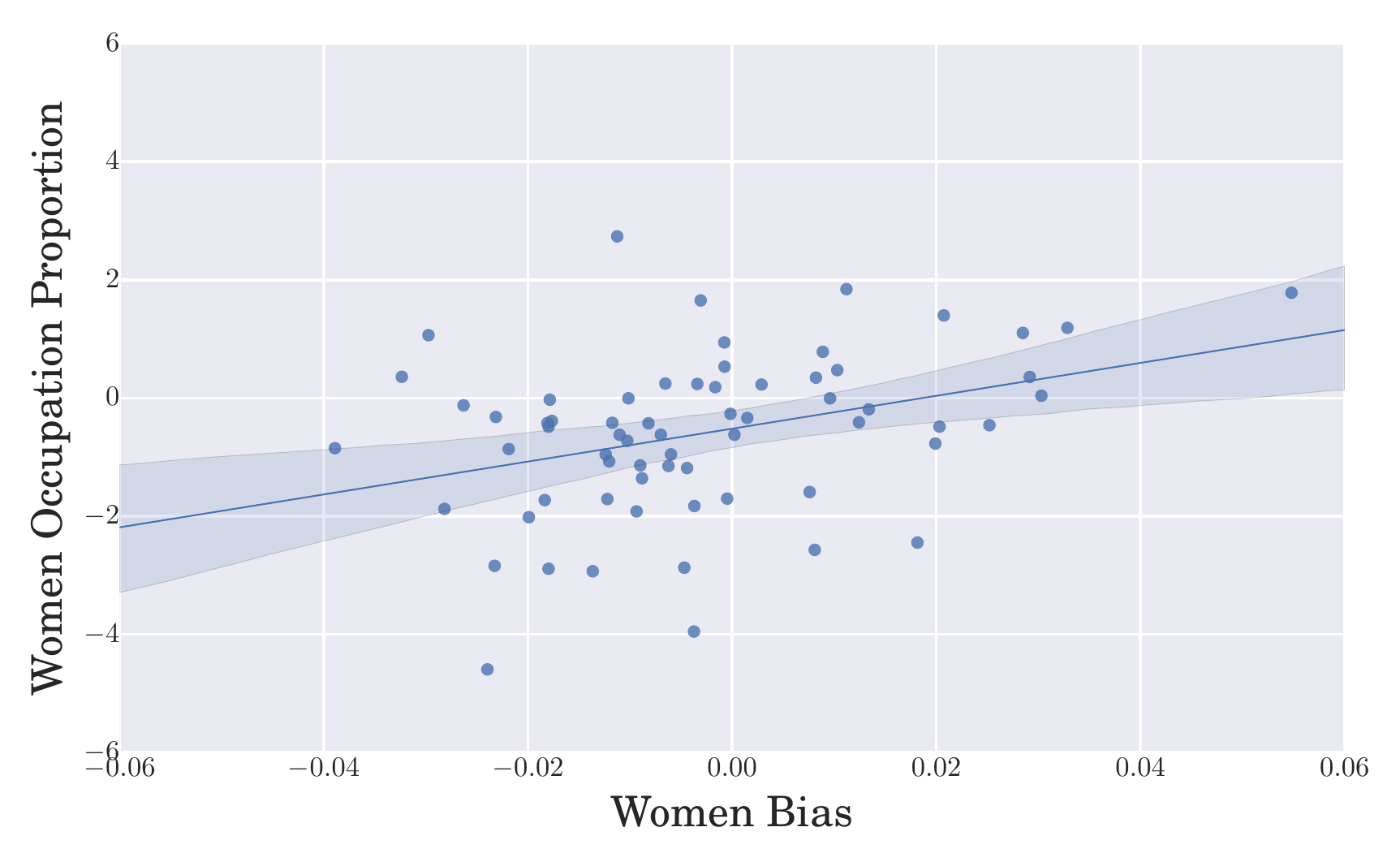}
		\caption{Common Crawl GloVe. $p< .005$, r-squared$=.127$.}
	\end{subfigure}
	\begin{subfigure}[H]{.45\textwidth}
		\includegraphics[width=\linewidth]{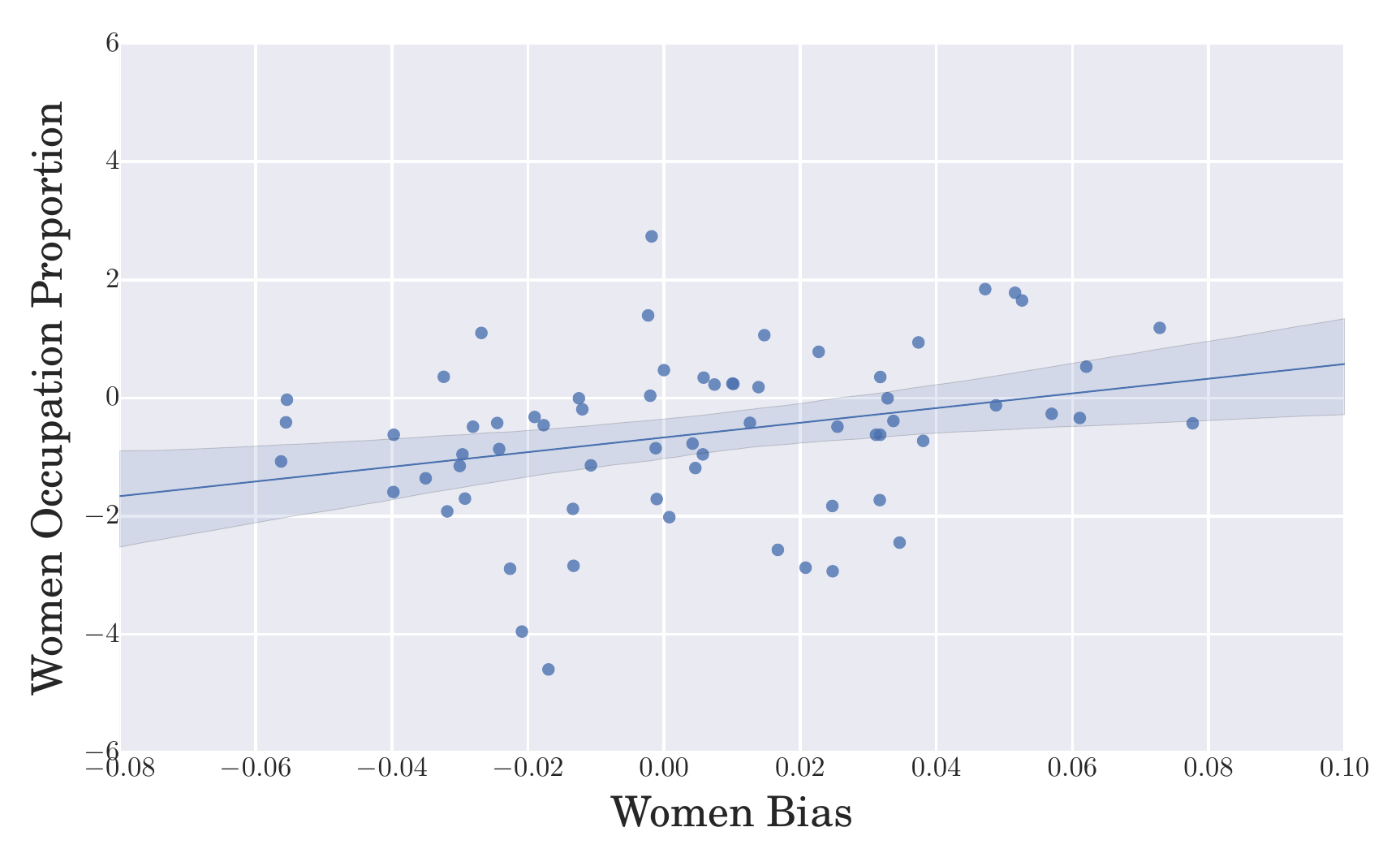}
		\caption{Wikipedia GloVe.  $p< .02$, r-squared$=.085$.}
	\end{subfigure}
	
	\caption{Percent woman in an occupation vs relative norm distance from occupations to the respective gender words in Common Crawl and Wikipedia. We note that for these datasets, there seems to be a slight systemic shift away from the origin, in contrast to the SGNS and Google News embeddings.}
	\label{fig:occ_percents_static_scatter_various}
\end{figure}

\begin{figure}[H]
	\centering
	
	\begin{subfigure}[H]{.7\linewidth}
		\includegraphics[width=\linewidth]{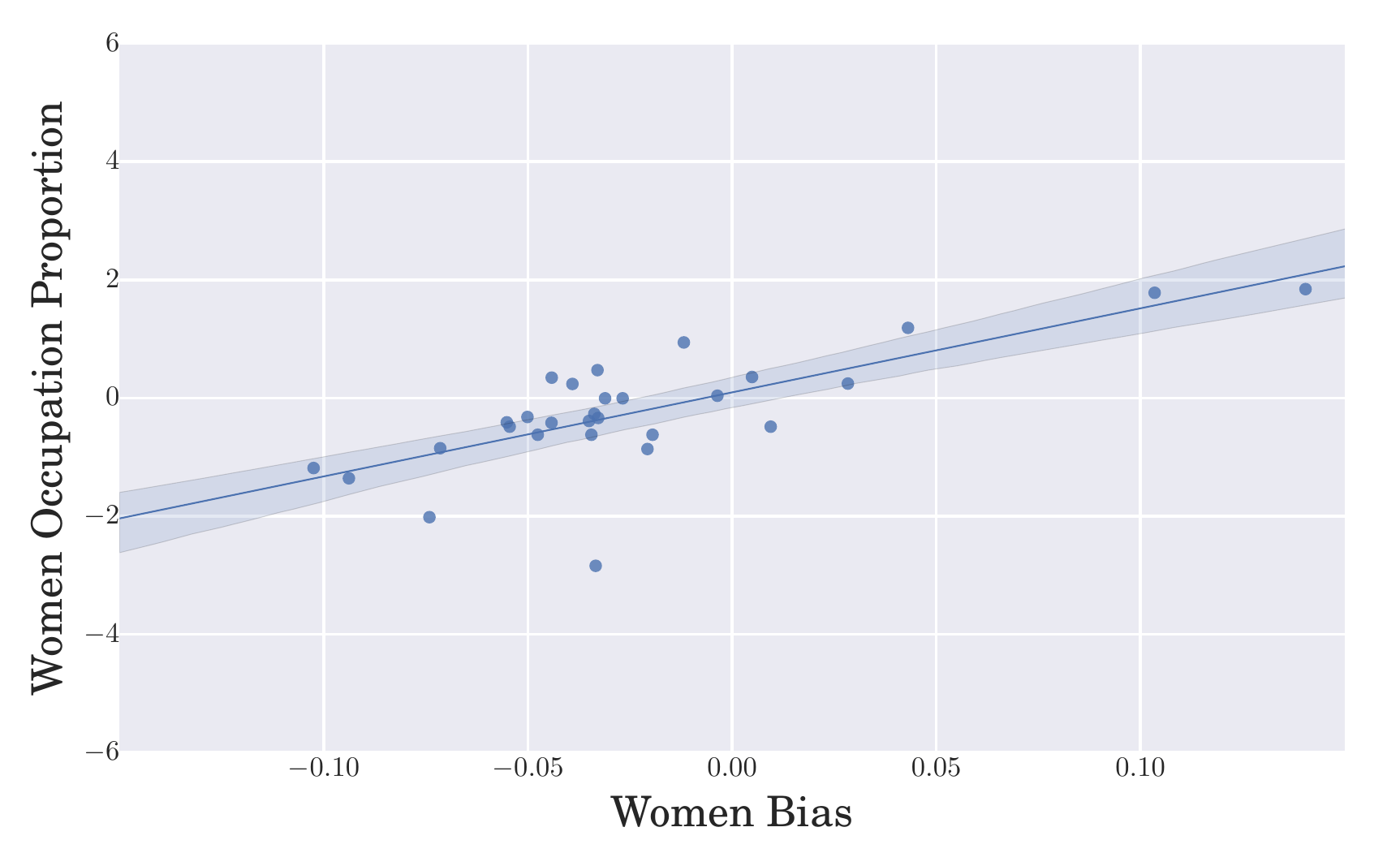}
	\end{subfigure}
	\caption{Percent woman in an occupation vs relative norm distance from \textit{professional} occupations in Google News vectors. $p< 10^{-5}$, r-squared$=.548$. Note that there is little difference in model from all occupations.}
	\label{fig:occ_percents_static_scatter_professional}
\end{figure}

\begin{table}[H]
	\centering
	\caption{Most Man and Woman \textit{residuals} when regressing embedding bias vs occupation gender proportion. The more Woman (Man) a residual the more biased the embedding is toward Women (Men) above what occupation proportions would suggest. }
	\label{tab:residualgender}
	\begin{tabular}{c|c}
		Man                 & Woman                          \\ \hline
		secretary            & nurse                     \\
		mechanic            & housekeeper                             \\
		musician                & gardener                      \\
		architect             & clerk                   \\
		janitor           & librarian                        \\
		carpenter              & sailor                     \\
		broker              & judge                         \\
		geologist             & artist                  \\
		accountant        & dancer              \\
		economist            & painter            
	\end{tabular}
\end{table}

\FloatBarrier
\subsection{Dynamic Analysis}
This section contains additional information for gender bias associated with the ``dynamic'' or over-time analysis. We first show the occupations and adjectives most associated with men and women, respectively, for each decade. Then we present additional information such as regression tables and plots with SVD embeddings for robustness.
\label{sec:genderappdynamic}

\subsubsection{Tables with top occupations and adjectives for each gender}
This subsection contains the top occupations and adjectives for each gender for each decade. We caution that due to the noisy nature of embeddings, these tables must be analyzed in the aggregate rather than focusing on individual associations.
\begin{table}[H]
	\begin{center}
		\makebox[\textwidth][c]{
			\begin{tabular}{ccccccccc}
				1910 & 1920 & 1930 & 1940 & 1950 & 1960 & 1970 & 1980 & 1990 \\\hline
				mathematician & accountant & engineer & surveyor & architect & lawyer & architect & architect & architect \\
				soldier & surveyor & architect & architect & engineer & architect & engineer & auctioneer & mathematician \\
				architect & architect & lawyer & engineer & mathematician & surveyor & judge & judge & surveyor \\
				surveyor & lawyer & surveyor & smith & lawyer & soldier & economist & surveyor & engineer \\
				administrator & mathematician & manager & sheriff & sheriff & engineer & soldier & sheriff & pilot \\
				lawyer & sheriff & pilot & lawyer & postmaster & pilot & author & author & lawyer \\
				judge & engineer & author & scientist & surveyor & scientist & surveyor & engineer & author \\
				scientist & statistician & scientist & author & scientist & economist & administrator & broker & judge \\
				author & mason & mathematician & economist & author & author & mason & inspector & soldier \\
				economist & scientist & accountant & mason & soldier & mason & mathematician & police & blacksmith
			\end{tabular}
		}
		\caption{Most Man occupations in each decade in the SGNS embeddings.}
	\end{center}
\end{table}
\begin{table}[H]
	\begin{center}
		\makebox[\textwidth][c]{
			\begin{tabular}{ccccccccc}
				1910 & 1920 & 1930 & 1940 & 1950 & 1960 & 1970 & 1980 & 1990 \\\hline
				nurse & nurse & nurse & nurse & nurse & nurse & nurse & nurse & nurse \\
				attendant & housekeeper & housekeeper & attendant & housekeeper & attendant & dancer & dancer & housekeeper \\
				housekeeper & attendant & attendant & janitor & attendant & dancer & housekeeper & attendant & midwife \\
				cashier & dancer & dancer & housekeeper & dancer & housekeeper & attendant & housekeeper & dentist \\
				cook & teacher & janitor & midwife & cook & photographer & conductor & midwife & student \\
				bailiff & supervisor & midwife & dentist & gardener & midwife & dentist & statistician & dancer \\
				porter & cook & clerical & cook & cashier & dentist & statistician & student & supervisor \\
				operator & doctor & dentist & clerical & midwife & janitor & baker & conductor & bailiff \\
				supervisor & dentist & cook & clergy & musician & cook & clerical & dentist & physician \\
				clergy & mechanic & teacher & sailor & sailor & porter & sailor & supervisor & doctor
			\end{tabular}
		}
		\caption{Most Woman occupations in each decade in the SGNS embeddings.}
	\end{center}
\end{table}
\begin{table}[H]
	\begin{center}
		\makebox[\textwidth][c]{
			\begin{tabular}{ccccccccc}
				1910 & 1920 & 1930 & 1940 & 1950 & 1960 & 1970 & 1980 & 1990 \\\hline
				honorable & regimental & honorable & honorable & knowledge & gallant & honorable & honorable & honorable \\
				gallant & honorable & trusting & conservative & gallant & honorable & wise & loyal & regimental \\
				regimental & stoic & courageous & ambitious & honorable & sage & knowledge & petty & unreliable \\
				skillful & political & gallant & shrewd & directed & regimental & gallant & gallant & skillful \\
				disobedient & sage & confident & regimental & regimental & knowledge & insulting & lyrical & gallant \\
				faithful & ambitious & adventurous & knowledge & efficient & wise & trusting & honest & honest \\
				wise & reserved & experimental & destructive & sage & conservative & honest & faithful & loyal \\
				obedient & progressive & efficient & misguided & wise & honest & providential & obedient & wise \\
				obnoxious & unprincipled & predatory & gallant & faithful & adventurous & modern & wise & directed \\
				steadfast & shrewd & modern & petty & creative & efficient & regimental & hostile & courageous
			\end{tabular}
		}
		\caption{Most Man adjectives in each decade in the SGNS embeddings.}
	\end{center}
\end{table}
\begin{table}[H]
	\begin{center}
		\makebox[\textwidth][c]{
			\begin{tabular}{ccccccccc}
				1910 & 1920 & 1930 & 1940 & 1950 & 1960 & 1970 & 1980 & 1990 \\\hline
				charming & charming & charming & delicate & delicate & sweet & attractive & maternal & maternal \\
				placid & relaxed & delicate & placid & sweet & charming & maternal & attractive & morbid \\
				delicate & delicate & soft & sweet & charming & soft & charming & masculine & artificial \\
				passionate & amiable & hysterical & gentle & transparent & relaxed & sweet & impassive & physical \\
				sweet & hysterical & transparent & soft & placid & attractive & caring & emotional & caring \\
				dreamy & placid & sweet & warm & childish & placid & venomous & protective & emotional \\
				indulgent & soft & relaxed & charming & soft & delicate & silly & relaxed & protective \\
				playful & gentle & shy & childish & colorless & maternal & neat & charming & attractive \\
				mellow & attractive & maternal & irritable & tasteless & indulgent & delicate & naive & soft \\
				sentimental & sweet & smooth & maternal & agreeable & gentle & sensitive & responsive & tidy
			\end{tabular}
		}
		\caption{Most Woman adjectives in each decade in the SGNS embeddings.}
		\label{tab:mostwomanadjectivesovertime}
	\end{center}
\end{table}

\subsubsection{Adjectives}
\FloatBarrier
This subsection contains the supplementary material related to adjectives. In particular, we show the scatter plots and fit information corresponding the stereotype scores from~\citep{williams_sex_1977} and~\citep{williams_measuring_1990} to the SGNS and SVD embeddings from the respective decades. 
\begin{figure}[H]
	\centering
	\begin{subfigure}[H]{.4\linewidth}
		\includegraphics[width=\linewidth]{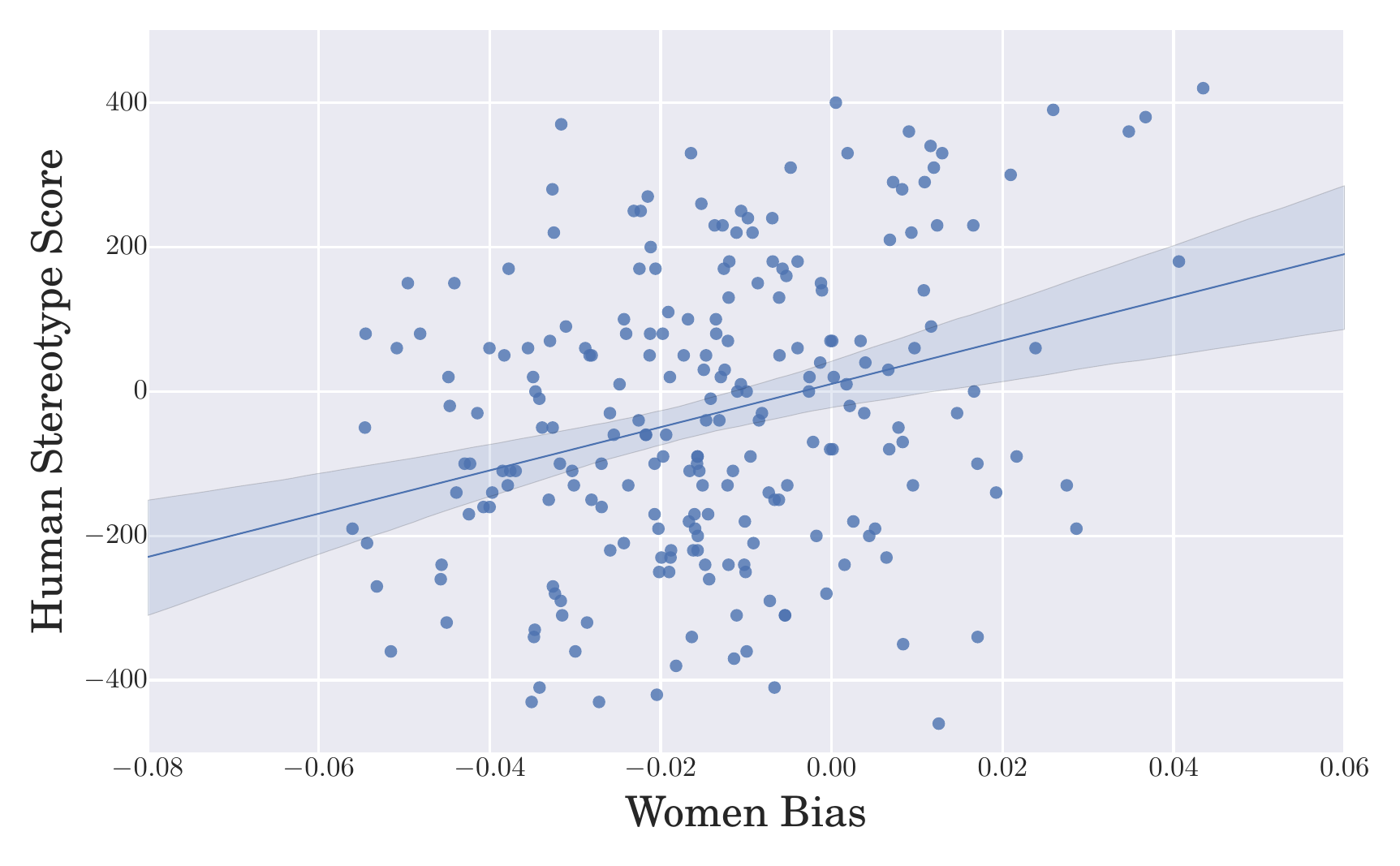}
		\caption{1990 SGNS vectors woman bias vs subjective scores from~\citep{williams_measuring_1990}. $p< 10^{-5}$, r-squared$=.086$.} 
	\end{subfigure}
	\begin{subfigure}[H]{.4\linewidth}
		\includegraphics[width=\linewidth]{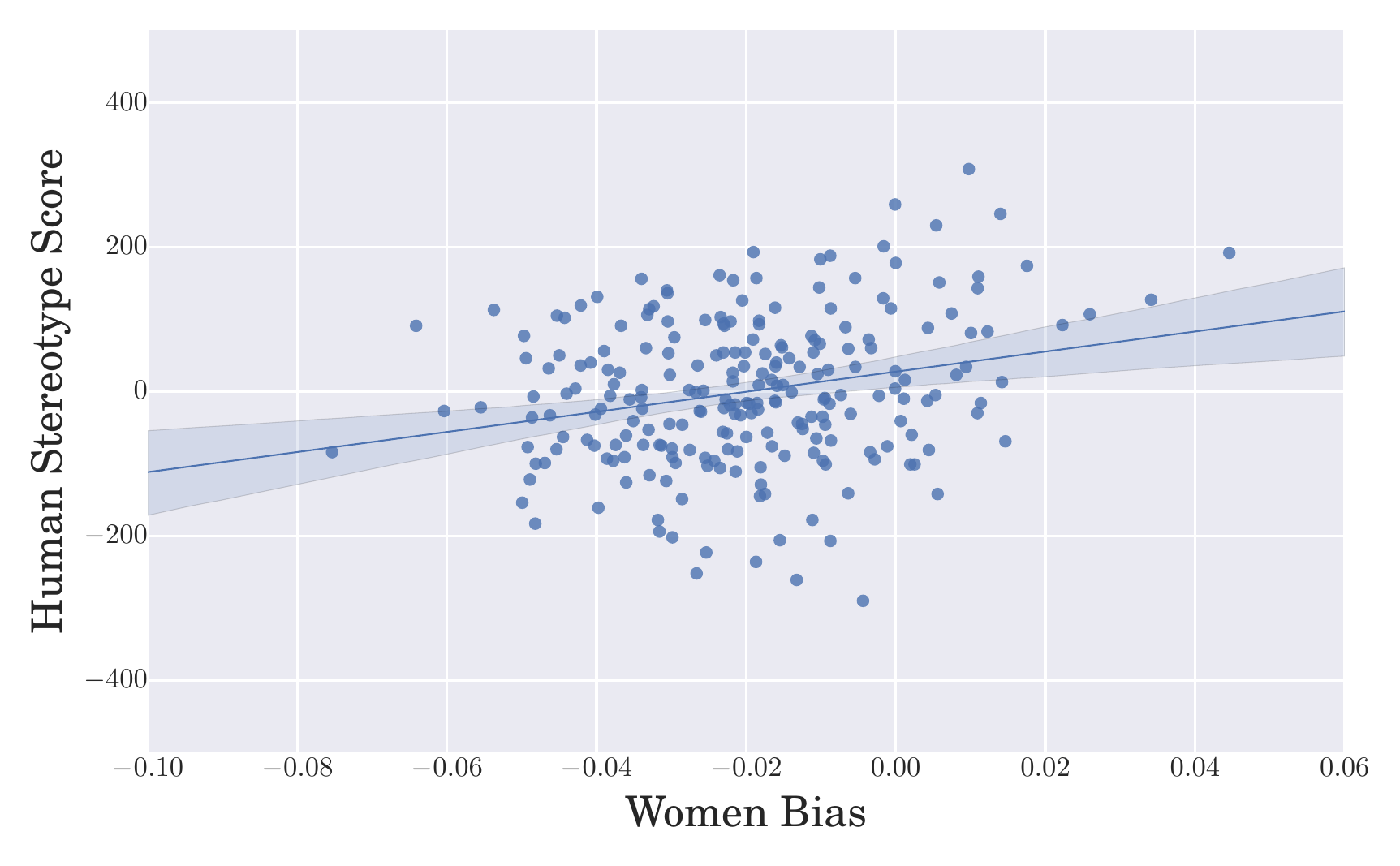}
		\caption{1970 SGNS vectors woman bias vs subjective scores from~\citep{williams_sex_1977}. $p< .0002$, r-squared$=.062$.}
	\end{subfigure}
	
	\begin{subfigure}[H]{.4\linewidth}
		\includegraphics[width=\linewidth]{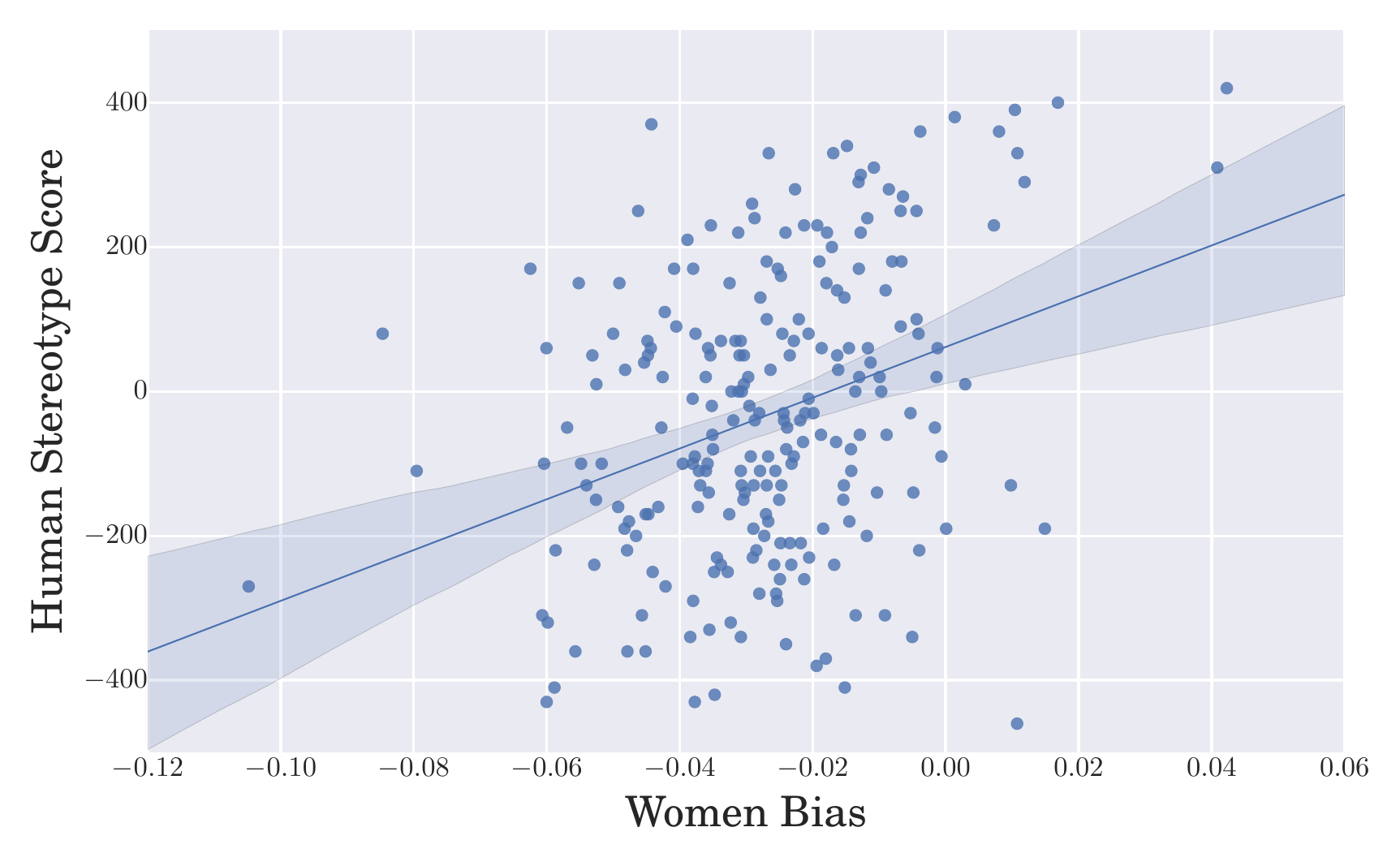}
		\caption{1990 SVD vectors woman bias vs subjective scores from~\citep{williams_measuring_1990}. $p< 10^{-6}$, r-squared$=.113$.} 
	\end{subfigure}
	\begin{subfigure}[H]{.4\linewidth}
		\includegraphics[width=\linewidth]{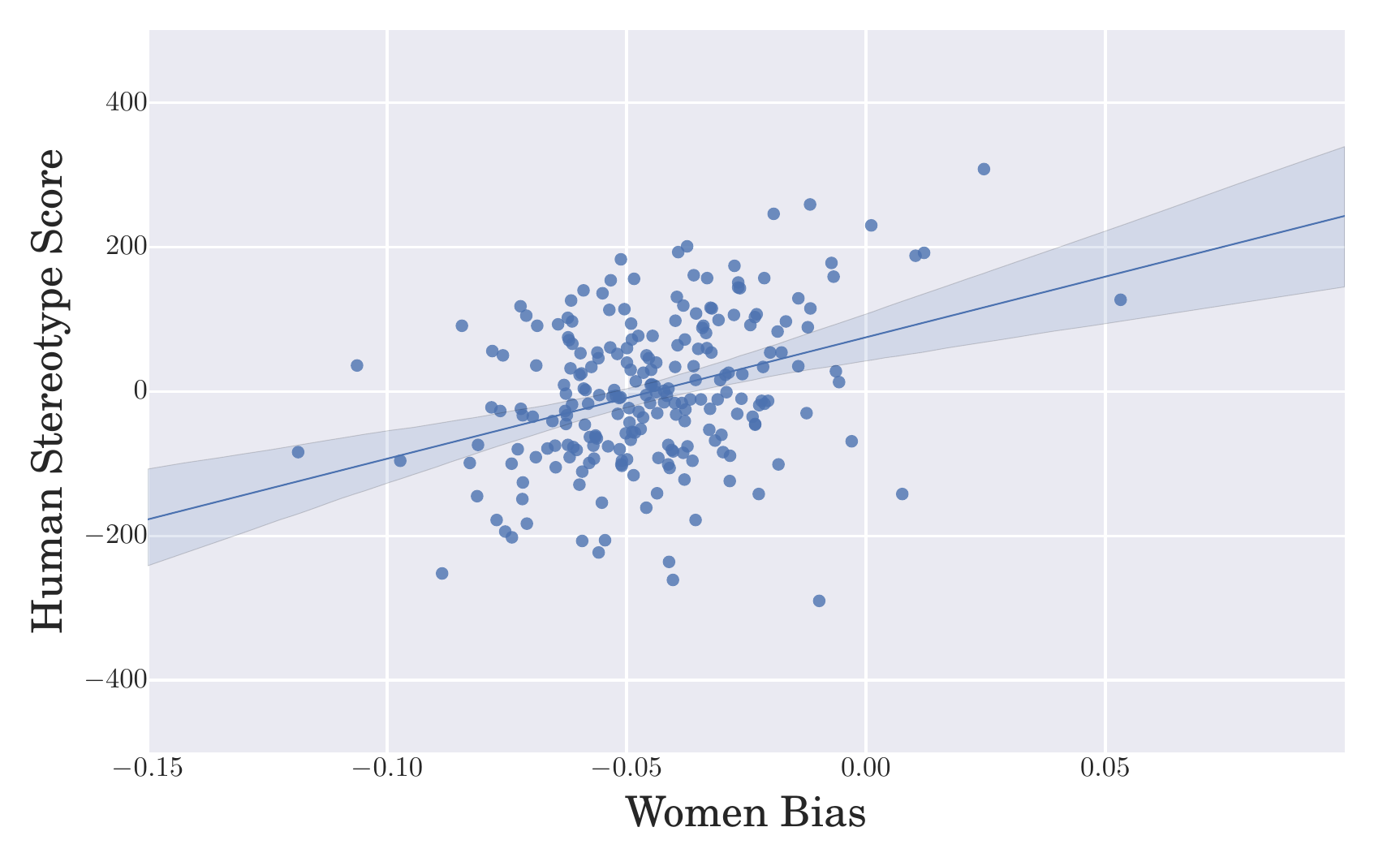}
		\caption{1970 SVD vectors woman bias vs subjective scores from~\citep{williams_sex_1977}. $p< 10^{-7}$, r-squared$=.127$.}
	\end{subfigure}
	\caption{Human stereotype score vs relative norm distance from adjectives to the respective gender words in 1970 and 1990, using both SGNS and SVD vectors}
	\label{fig:williamsbest_static_scatter_various}
\end{figure}

Note the stereotype scores used in the plots are linear transformations of the scores reported in the original papers, done to standardize figures. Given a score $r$ from 1990~\citep{williams_measuring_1990}, the transformed score $r' = 500 - 10r$. Given a score from 1970~\citep{williams_sex_1977}, the transformed score $r' = 500 - r$. 

\subsubsection{Supplementary for gender time dynamical analysis, SGNS embeddings}

Here, we provide regression tables and compare models trained for each year to the single model trained over all years.

\begin{table}[H]
	\begin{center}
		\begin{tabular}{lclc}
			\toprule
			\textbf{Dep. Variable:}             & Woman Occupation Proportion & \textbf{  R-squared:         } &     0.235   \\
			\textbf{Model:}                     &             OLS              & \textbf{  Adj. R-squared:    } &     0.233   \\
			\textbf{Method:}                    &        Least Squares         & \textbf{  F-statistic:       } &     190.4   \\
			\textbf{Date:}                      &       Fri, 08 Sep 2017       & \textbf{  Prob (F-statistic):} &  5.71e-38   \\
			\textbf{Time:}                      &           14:16:19           & \textbf{  Log-Likelihood:    } &   -1271.6   \\
			\textbf{No. Observations:}          &               623            & \textbf{  AIC:               } &     2547.   \\
			\textbf{Df Residuals:}              &               621            & \textbf{  BIC:               } &     2556.   \\
			\textbf{Df Model:}                  &                 1            & \textbf{                     } &             \\
			\bottomrule
		\end{tabular}
		\begin{tabular}{lccccc}
			& \textbf{coef} & \textbf{std err} & \textbf{t} & \textbf{P$>$$|$t$|$} & \textbf{[95.0\% Conf. Int.]}  \\
			\midrule
			\textbf{Relative Woman Similarity} &      36.0112  &        2.610     &    13.798  &         0.000        &        30.886    41.136       \\
			\textbf{const}                      &       0.0561  &        0.149     &     0.377  &         0.706        &        -0.236     0.348       \\
			\bottomrule
		\end{tabular}
		\begin{tabular}{lclc}
			\textbf{Omnibus:}       &  2.527 & \textbf{  Durbin-Watson:     } &    2.064  \\
			\textbf{Prob(Omnibus):} &  0.283 & \textbf{  Jarque-Bera (JB):  } &    2.603  \\
			\textbf{Skew:}          & -0.048 & \textbf{  Prob(JB):          } &    0.272  \\
			\textbf{Kurtosis:}      &  3.302 & \textbf{  Cond. No.          } &     35.0  \\
			\bottomrule
		\end{tabular}
		\caption{Regression table for Figure~\ref{fig:occ_percents_static_scatter_allyears}.}
	\end{center}
\end{table}

\begin{table}[H]
	\begin{center}
		\begin{tabular}{lclc}
			\toprule
			\textbf{Dep. Variable:}    & Woman Occupation Proportion & \textbf{  R-squared:         } &     0.072   \\
			\textbf{Model:}            &             OLS              & \textbf{  Adj. R-squared:    } &     0.060   \\
			\textbf{Method:}           &        Least Squares         & \textbf{  F-statistic:       } &     5.972   \\
			\textbf{Date:}             &       Sun, 10 Sep 2017       & \textbf{  Prob (F-statistic):} &  2.15e-07   \\
			\textbf{Time:}             &           14:18:53           & \textbf{  Log-Likelihood:    } &   -1331.6   \\
			\textbf{No. Observations:} &               623            & \textbf{  AIC:               } &     2681.   \\
			\textbf{Df Residuals:}     &               614            & \textbf{  BIC:               } &     2721.   \\
			\textbf{Df Model:}         &                 8            & \textbf{                     } &             \\
			\bottomrule
		\end{tabular}
		\begin{tabular}{lccccc}
			& \textbf{coef} & \textbf{std err} & \textbf{t} & \textbf{P$>$$|$t$|$} & \textbf{[95.0\% Conf. Int.]}  \\
			\midrule
			\textbf{const}     &      -1.5611  &        0.075     &   -20.923  &         0.000        &        -1.708    -1.415       \\
			\textbf{yr-1910.0} &      -0.9255  &        0.241     &    -3.839  &         0.000        &        -1.399    -0.452       \\
			\textbf{yr-1920.0} &      -0.6298  &        0.238     &    -2.648  &         0.008        &        -1.097    -0.163       \\
			\textbf{yr-1930.0} &      -0.5559  &        0.227     &    -2.444  &         0.015        &        -1.003    -0.109       \\
			\textbf{yr-1940.0} &      -0.7718  &        0.246     &    -3.134  &         0.002        &        -1.255    -0.288       \\
			\textbf{yr-1950.0} &      -0.2511  &        0.232     &    -1.084  &         0.279        &        -0.706     0.204       \\
			\textbf{yr-1960.0} &      -0.0236  &        0.230     &    -0.103  &         0.918        &        -0.476     0.428       \\
			\textbf{yr-1970.0} &       0.2040  &        0.230     &     0.886  &         0.376        &        -0.248     0.656       \\
			\textbf{yr-1980.0} &       0.5712  &        0.233     &     2.450  &         0.015        &         0.113     1.029       \\
			\textbf{yr-1990.0} &       0.8214  &        0.233     &     3.523  &         0.000        &         0.363     1.279       \\
			\bottomrule
		\end{tabular}
		\caption{Regression for occupation proportion with the decade as a categorical independent variable. This table formalizes the understanding that average Woman occupation proportion has changed considerable over time.}
	\end{center}
\end{table}

\begin{figure}[H]
	\centering
	
	\begin{subfigure}[H]{.7\linewidth}
		\includegraphics[width=\linewidth]{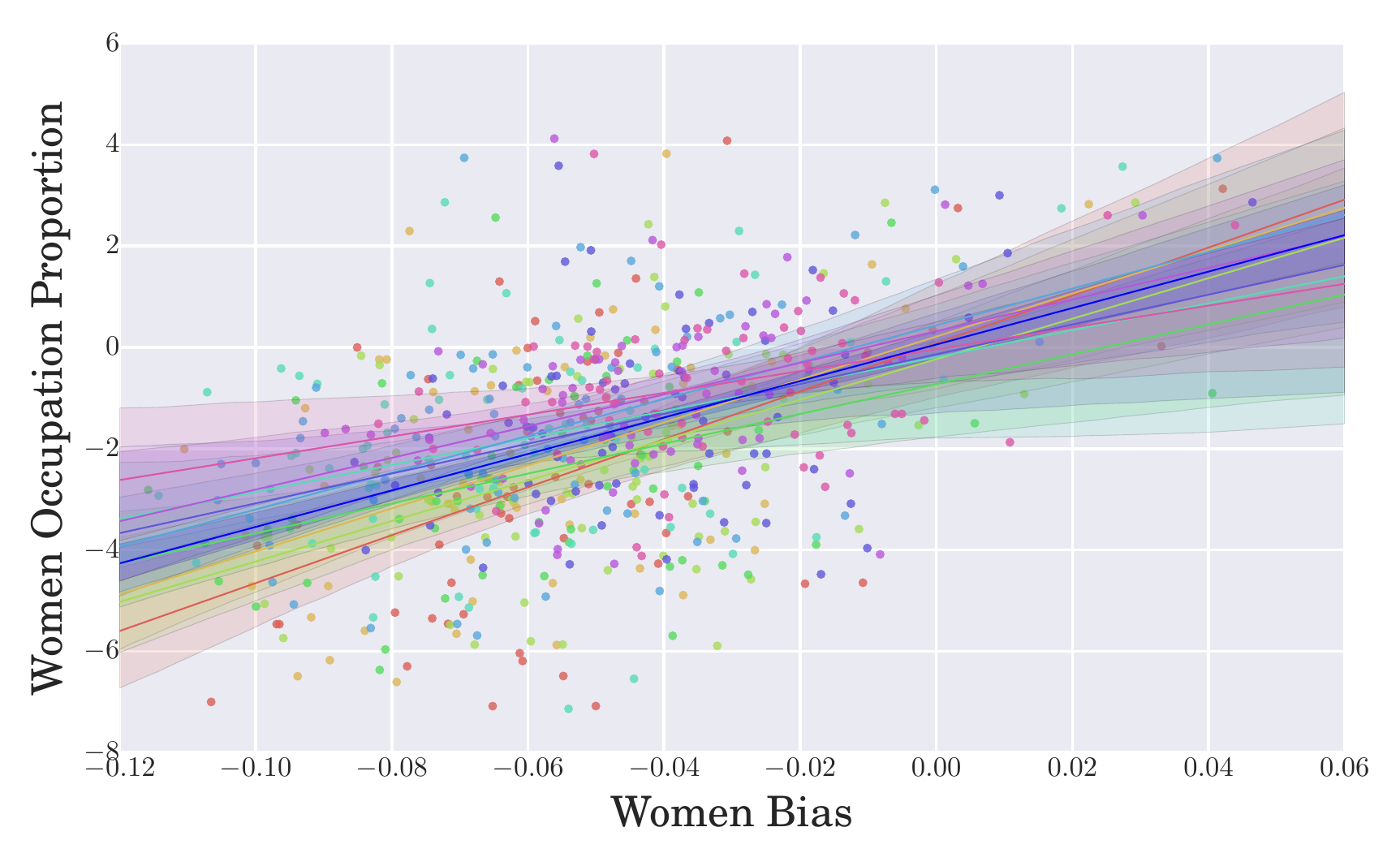}
	\end{subfigure}
	\caption{Scatter plot for occupation proportion vs embedding bias all years of SGNS embeddings, with individually trained models. Note that the regression coefficients (in Table~\ref{tab:overtimecoefficients}) are similar, especially for adjacent decades, and there is little loss training a single model for all years as opposed to individual per-year models.}
	\label{fig:occ_percents_static_scatter_allyears}
\end{figure}

\begin{table}[H]
	\centering
	\caption{Regression coefficients for each individually trained model.}
	\label{tab:overtimecoefficients}
	\begin{tabular}{c|cc}
		Year & coefficient $\pm$ std & p-value \\ \hline
		1910	&    $47.32  \pm 9.86$        &     0.000010 \\
		1920	&    $42.42  \pm 8.03$        &     0.000002 \\
		1930	&    $39.95  \pm 7.48$        &     0.000001 \\
		1940	&    $29.46  \pm 7.29$        &     0.000152 \\
		1950	&    $26.62  \pm 8.20$        &     0.0018 \\
		1960	&    $36.35  \pm 7.82$        &     0.000015 \\
		1970	&    $29.43  \pm 8.75$        &     0.00125 \\
		1980	&    $31.27  \pm 8.09$        &     0.00025 \\
		1990	&     $21.52  \pm 7.52$        &     0.0056   
	\end{tabular}
\end{table}
\begin{table}[H]
	\centering
	\caption{Model performance (occupation proportion vs embedding bias) for individual models trained on each decade compared to a single model trained for all decades. Note that the r-squared value for the combined model is $.235$.}
	\label{tab:overtimeMSE}
	\begin{tabular}{c|c|cc}
		Year & r-squared with individual model & MSE with individual model & MSE with combined model \\ \hline
		1910 & 0.268 & 4.73 & 5.18 \\
		1920 & 0.301 & 3.85 & 3.93 \\
		1930 & 0.284 & 3.51 & 3.75 \\
		1940 & 0.214 & 3.31 & 3.53 \\
		1950 & 0.133 & 3.77 & 3.95 \\
		1960 & 0.236 & 3.41 & 3.54 \\
		1970 & 0.139 & 2.89 & 2.92 \\
		1980 & 0.180 & 2.17 & 2.40 \\
		1990 & 0.107 & 1.94 & 2.19 
	\end{tabular}
\end{table}

\begin{figure}[H]
	\centering
	\includegraphics[width=.6\linewidth]{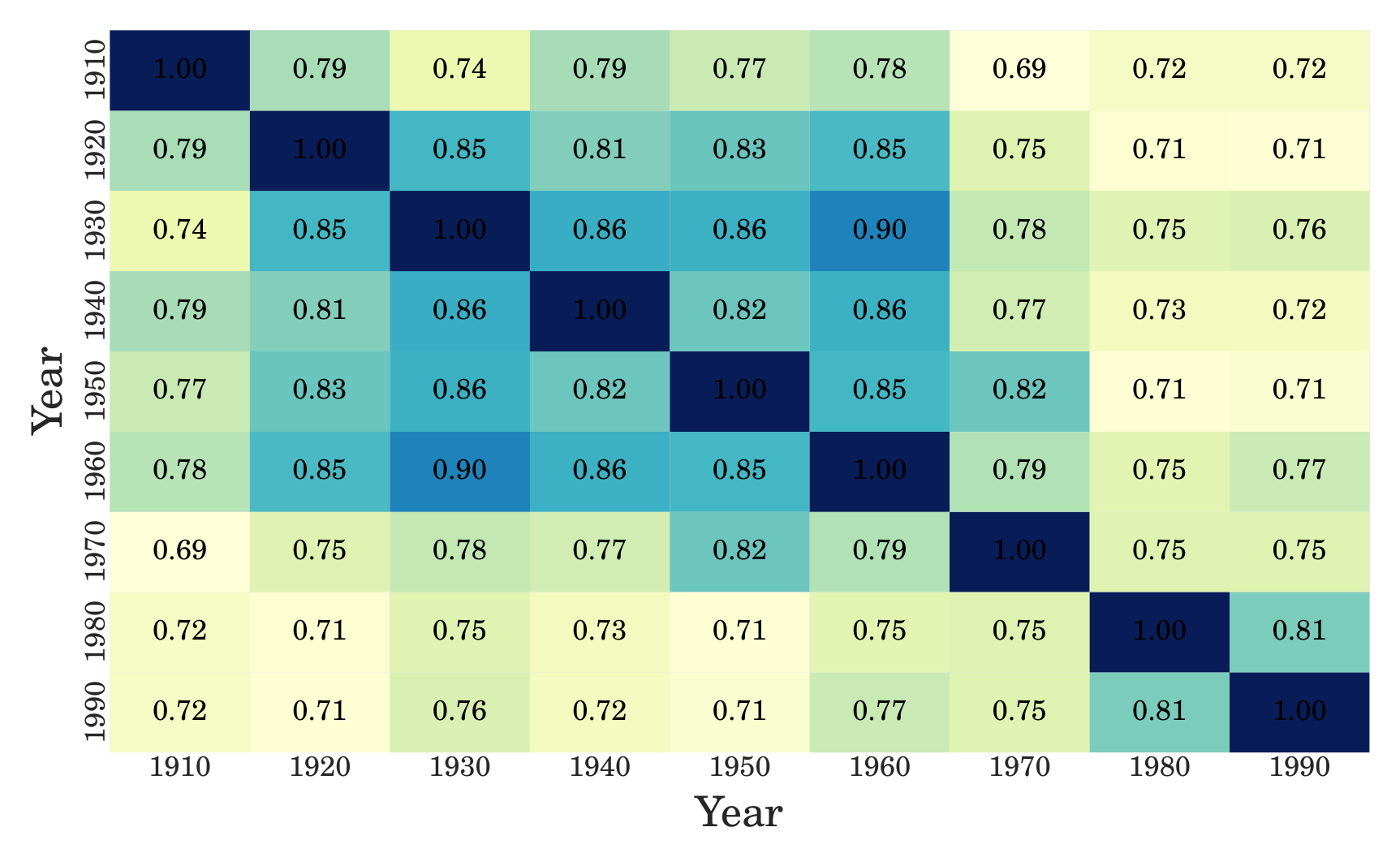}
	\caption{Pearson correlation in embedding bias scores for occupations over time between embeddings for each decade.} 
\end{figure}

\subsubsection{Supplementary for gender time dynamical analysis, SVD embeddings}
In this section we repeat entire gender time dynamical analysis, using SVD embeddings, as a robustness check.

\begin{table}[H]
	\begin{center}
		\begin{tabular}{lclc}
			\toprule
			\textbf{Dep. Variable:}             & Woman Occupation Proportion & \textbf{  R-squared:         } &     0.266   \\
			\textbf{Model:}                     &             OLS              & \textbf{  Adj. R-squared:    } &     0.265   \\
			\textbf{Method:}                    &        Least Squares         & \textbf{  F-statistic:       } &     225.1   \\
			\textbf{Date:}                      &       Sun, 10 Sep 2017       & \textbf{  Prob (F-statistic):} &  1.23e-43   \\
			\textbf{Time:}                      &           12:31:40           & \textbf{  Log-Likelihood:    } &   -1258.6   \\
			\textbf{No. Observations:}          &               623            & \textbf{  AIC:               } &     2521.   \\
			\textbf{Df Residuals:}              &               621            & \textbf{  BIC:               } &     2530.   \\
			\textbf{Df Model:}                  &                 1            & \textbf{                     } &             \\
			\bottomrule
		\end{tabular}
		\begin{tabular}{lccccc}
			& \textbf{coef} & \textbf{std err} & \textbf{t} & \textbf{P$>$$|$t$|$} & \textbf{[95.0\% Conf. Int.]}  \\
			\midrule
			\textbf{Relative Woman Similarity} &      33.3034  &        2.220     &    15.002  &         0.000        &        28.944    37.663       \\
			\textbf{const}                      &       0.7163  &        0.178     &     4.022  &         0.000        &         0.367     1.066       \\
			\bottomrule
		\end{tabular}
		\begin{tabular}{lclc}
			\textbf{Omnibus:}       &  7.552 & \textbf{  Durbin-Watson:     } &    1.920  \\
			\textbf{Prob(Omnibus):} &  0.023 & \textbf{  Jarque-Bera (JB):  } &   11.040  \\
			\textbf{Skew:}          &  0.031 & \textbf{  Prob(JB):          } &  0.00401  \\
			\textbf{Kurtosis:}      &  3.649 & \textbf{  Cond. No.          } &     30.5  \\
			\bottomrule
		\end{tabular}
		\caption{Regression table for Figure~\ref{fig:occ_percents_static_scatter_allyears_svd_singlemodel}}
	\end{center}
\end{table}

\begin{figure}[H]
	\centering
	\begin{subfigure}[H]{.45\linewidth}
		\includegraphics[width=\linewidth]{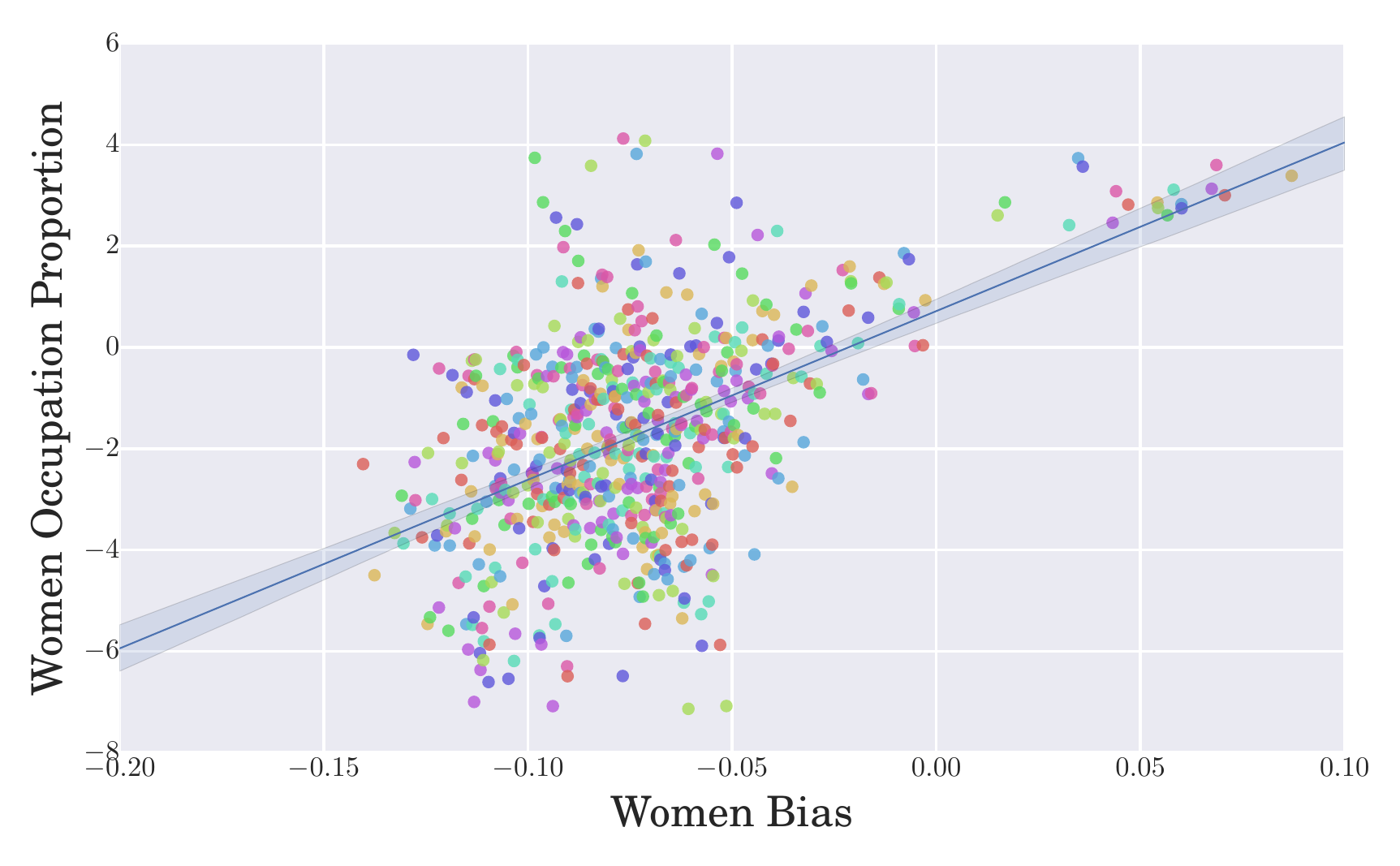}
		\caption{Single model trained for all time}
		\label{fig:occ_percents_static_scatter_allyears_svd_singlemodel}
	\end{subfigure}
	\begin{subfigure}[H]{.45\linewidth}
		\includegraphics[width=\linewidth]{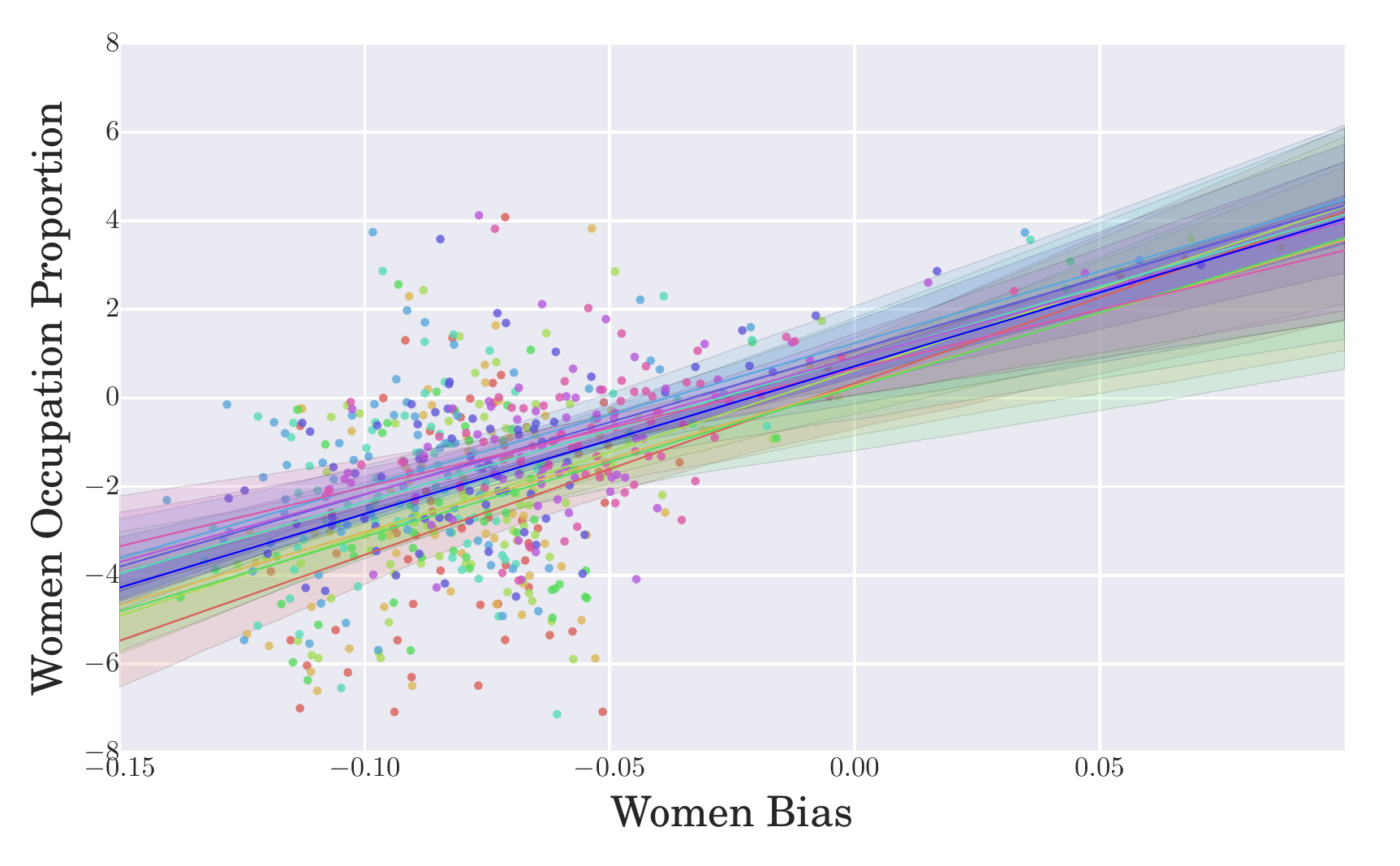}
		\caption{Separate model trained for each decade}
	\end{subfigure}
	\caption{Scatter plot for occupation proportion vs embedding bias all years of SVD embeddings, along with individually trained regression lines.}
	\label{fig:occ_percents_static_scatter_allyears_svd}
\end{figure}

\begin{table}[H]
	\centering
	\caption{Regression coefficients for each individually trained model.}
	\label{tab:overtimecoefficients_svd}
	\begin{tabular}{c|cc}
		Year & coefficient $\pm$ std & p-value \\ \hline
		1910	&    $38.72  \pm 8.61$        &     0.000030 \\
		1920	&    $32.96  \pm 7.00$        &     0.000014 \\
		1930	&    $36.96  \pm 6.91$        &      9.95e-07 \\
		1940	&    $33.73  \pm 6.89$        &     0.000008 \\
		1950	&    $32.40  \pm 6.84$        &     0.000011 \\
		1960	&    $32.36  \pm 6.71$        &     0.000008 \\
		1970	&    $32.60  \pm 5.74$        &     2.84e-07 \\
		1980	&    $30.71  \pm 6.48$        &     0.000012 \\
		1990	&     $26.71  \pm 5.81$        &     0.000019   
	\end{tabular}
\end{table}
\begin{table}[H]
	\centering
	\caption{Model performance (occupation proportion vs embedding bias) for individual models trained on each decade compared to a single model trained for all decades. Note that the r-squared value for the combined model is $.266$.}
	\label{tab:overtimepvaluess_svd}
	\begin{tabular}{c|c|cc}
		Year & r-squared with individual model & MSE with individual model & MSE with combined model \\ \hline
		1910 & 0.243 & 4.89 & 5.52 \\
		1920 & 0.254 & 4.10 & 4.28 \\
		1930 & 0.284 & 3.51 & 3.65 \\
		1940 & 0.285 & 3.01 & 3.27 \\
		1950 & 0.246 & 3.28 & 3.35 \\
		1960 & 0.249 & 3.35 & 3.72 \\
		1970 & 0.316 & 2.30 & 2.47 \\
		1980 & 0.248 & 1.99 & 2.12 \\
		1990 & 0.237 & 1.66 & 1.77 
	\end{tabular}
\end{table}

\begin{figure}[H]
	\centering
	\includegraphics[width=\linewidth]{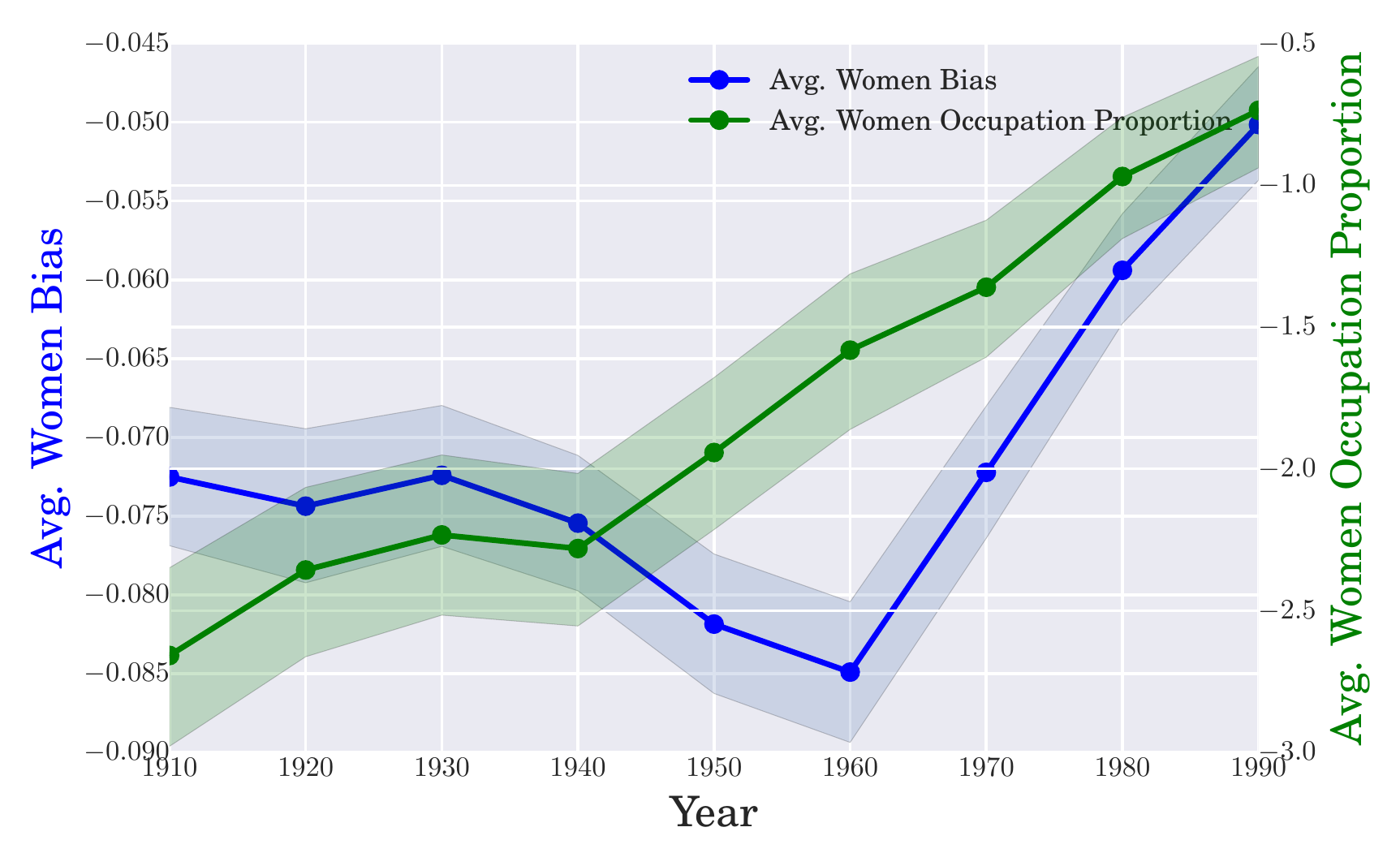}
	\caption{Gender bias over time in COHA dataset in occupations vs the average log proportion. In blue is the woman bias in the SVD embeddings, while the in green is the average log proportion of women in the occupation.}
	\label{}
\end{figure}

\begin{figure}[H]
	\centering	
	\begin{subfigure}[H]{.6\linewidth}
		\includegraphics[width=\linewidth]{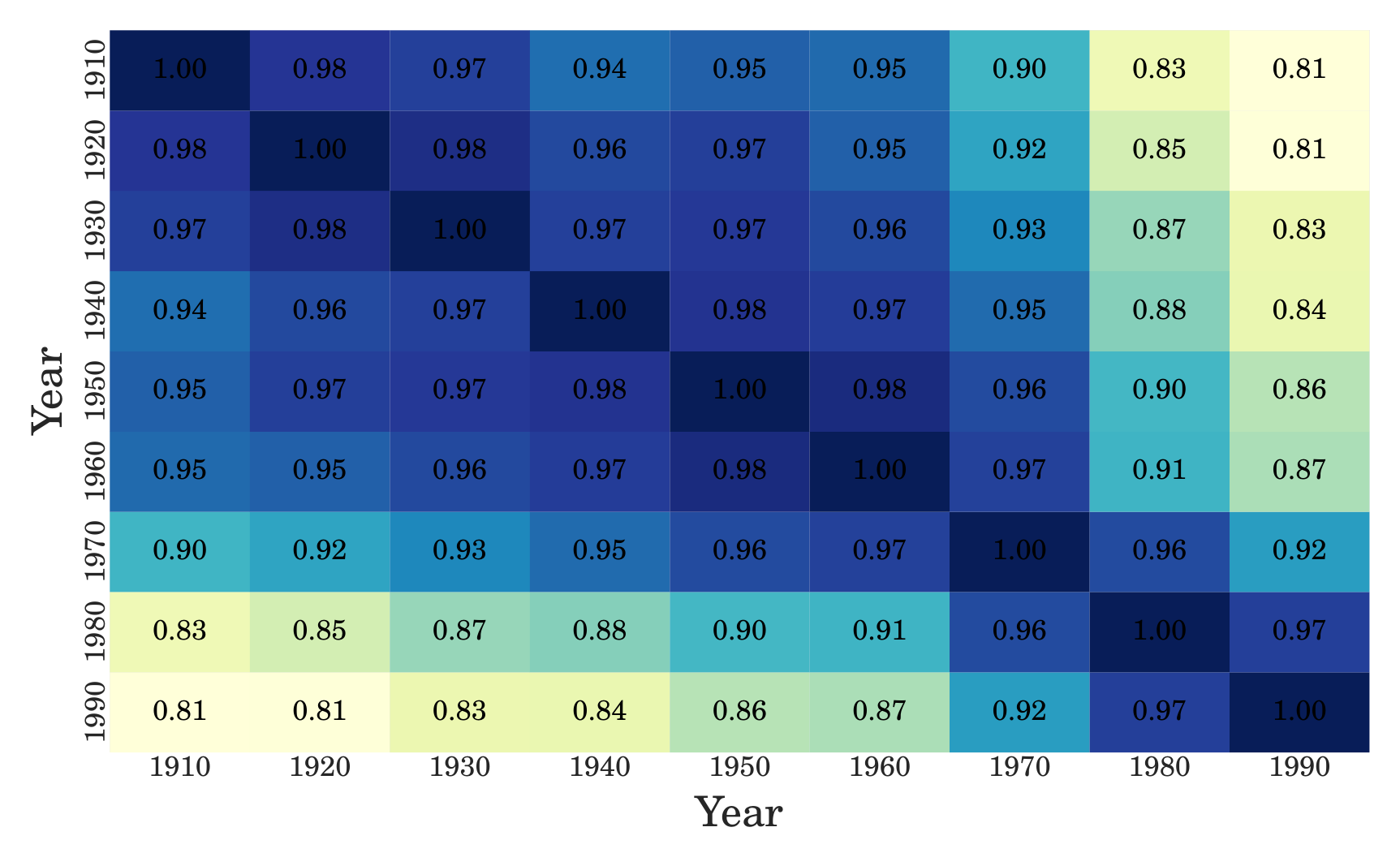}
		\caption{Pearson correlation in SVD embedding bias scores for occupations over time between embeddings for each decade.} 
	\end{subfigure}
	
	\begin{subfigure}[H]{.6\linewidth}
		\includegraphics[width=\linewidth]{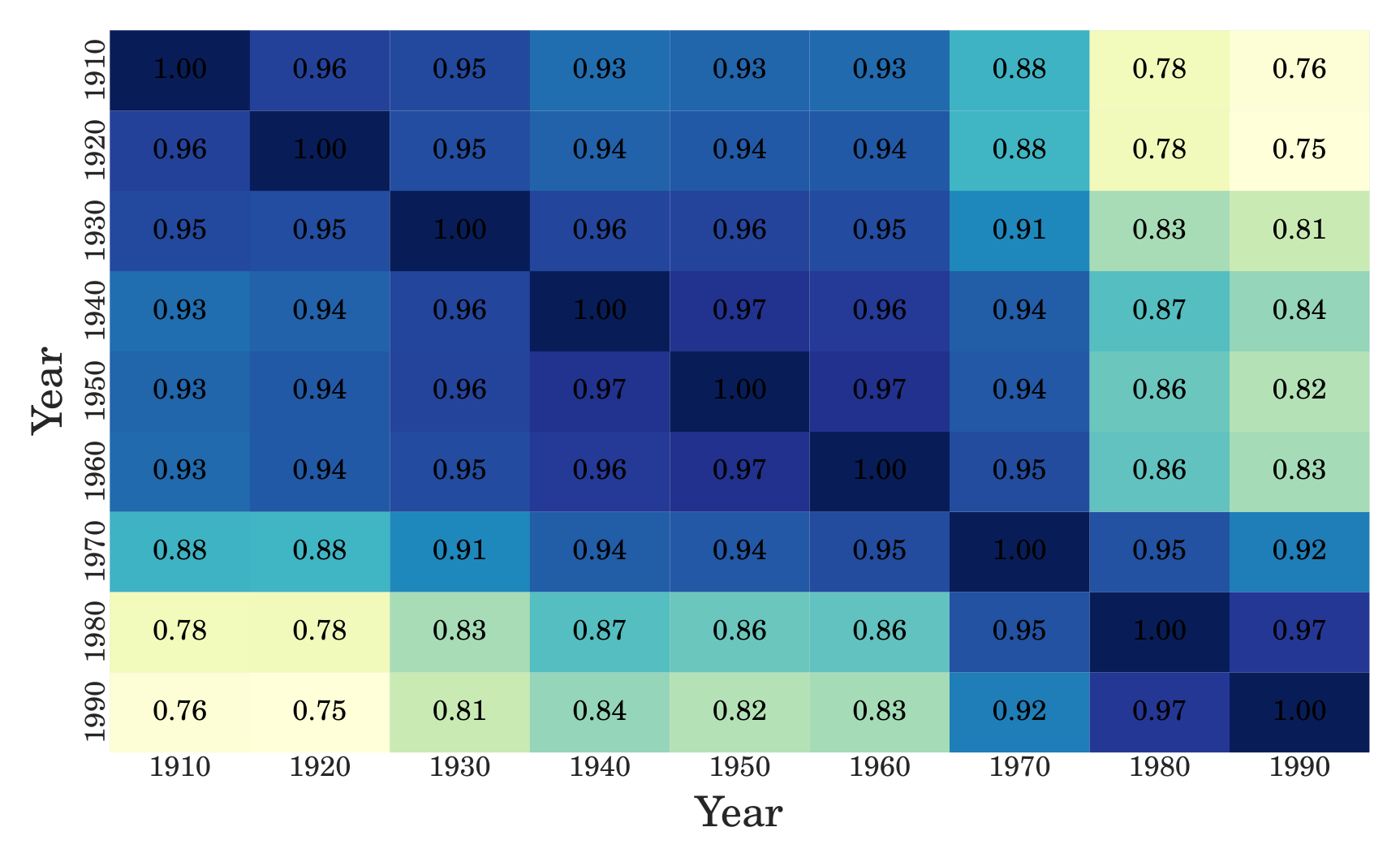}
		\caption{Pearson correlation in SVD embedding bias scores for adjectives over time between embeddings for each decade.} 	\end{subfigure}
	\label{}
\end{figure}
\section{Ethnic groups}

This section provides additional information and figures related to Section~\ref{sec:race}

\subsection{Snapshot Analysis}
\label{sec:raceappstatic}
This section contains supplementary material associated with ``snapshot'' analysis of ethnic bias.

\begin{figure}[H]
	\centering
	\begin{subfigure}[H]{.7\linewidth}
		\includegraphics[width=\linewidth]{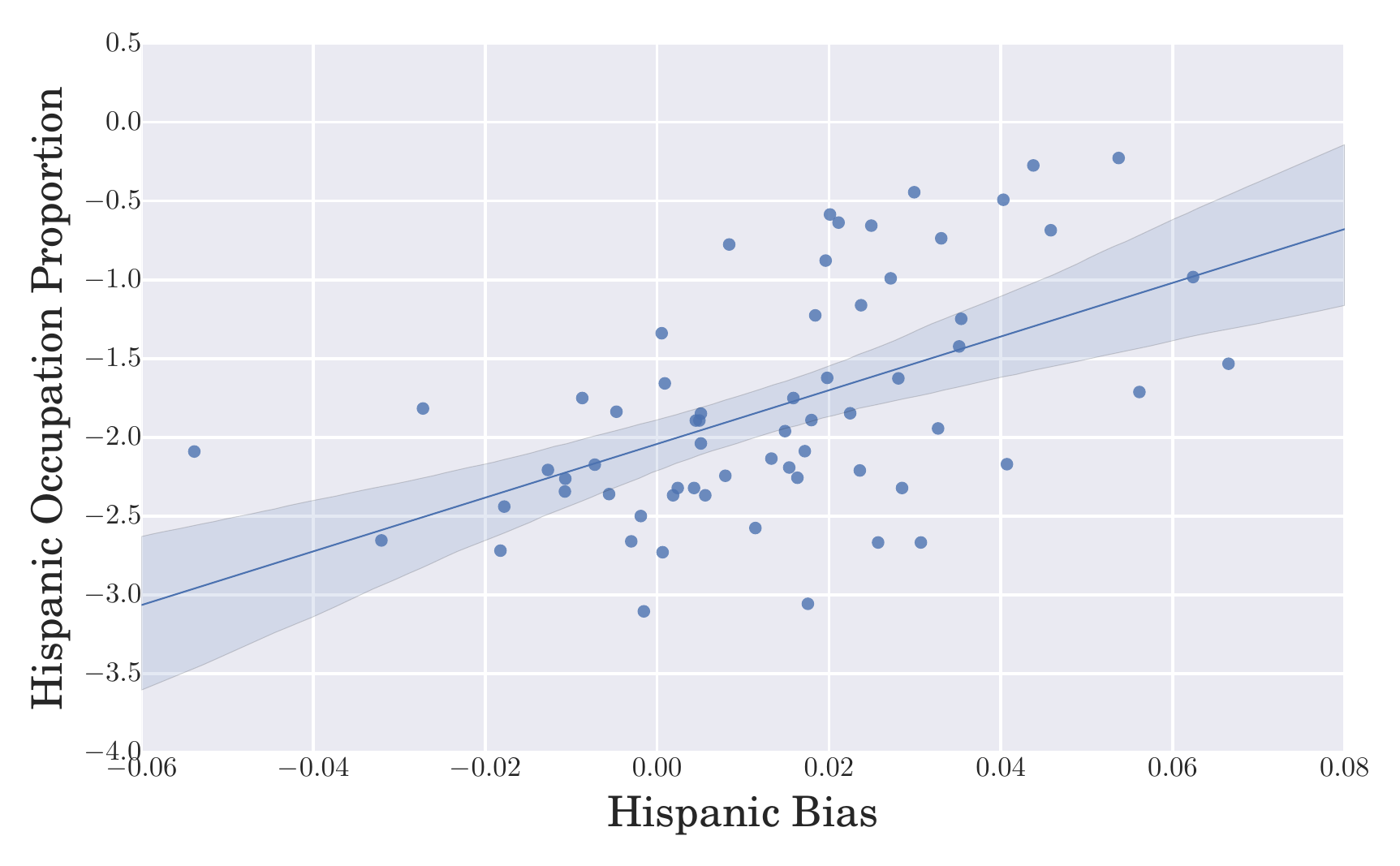}
		\caption{Conditional Log Proportion of Hispanics (compared to Whites) in an occupation vs relative norm distance from occupations to the respective gender words in Google News vectors. More positive indicates more Hispanic associated, for both proportion of occupations and for relative distance. $p < 10^{-5}$ and with r-squared$ = .277$.} 
		\label{fig:appscatterocchispanics}
	\end{subfigure}
	
	\begin{subfigure}[H]{.7\linewidth}
		\includegraphics[width=\linewidth]{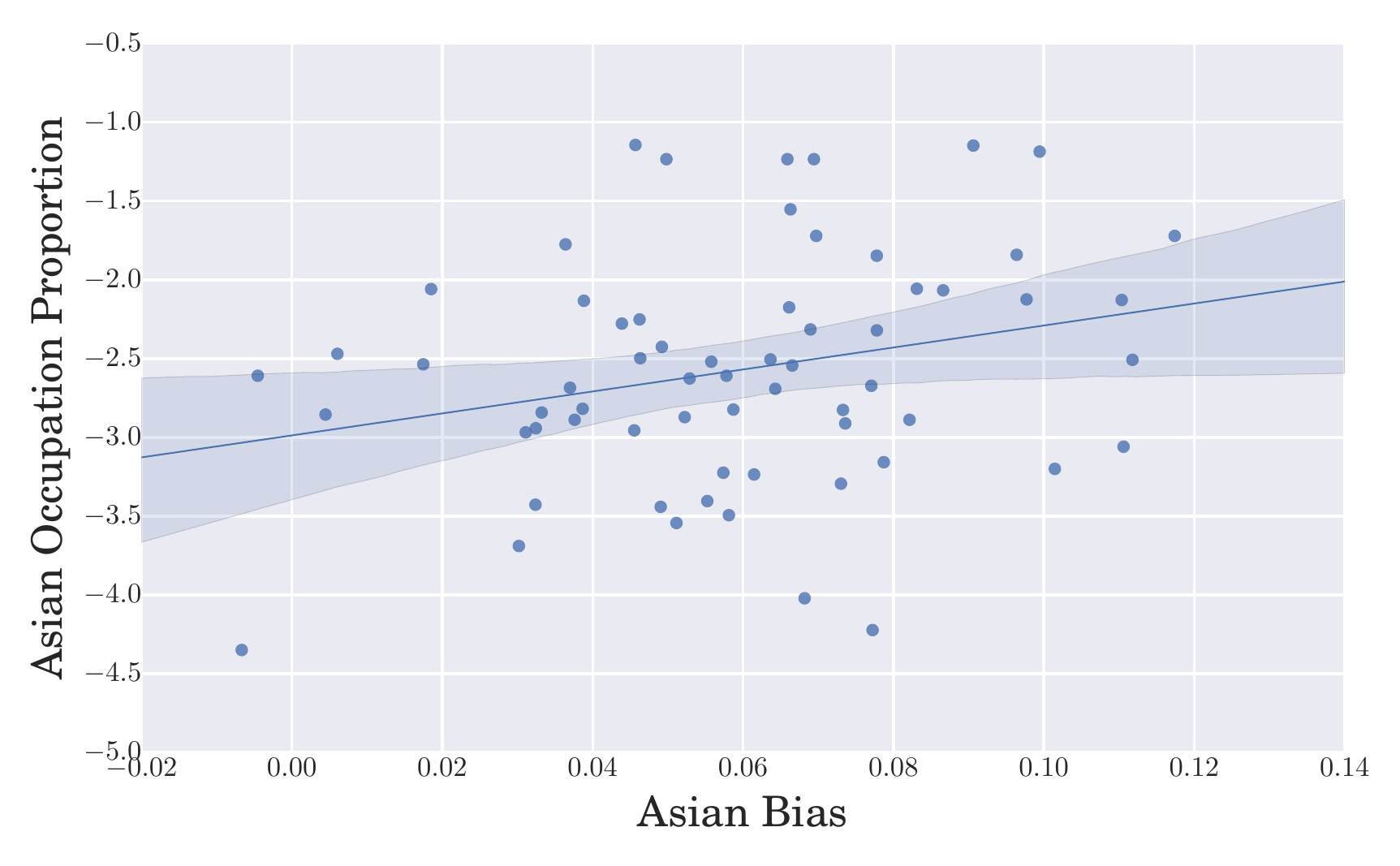}
		\caption{Conditional Log Proportion of Asians (compared to Whites) in an occupation vs relative norm distance from occupations to the respective gender words in Google News vectors. More positive indicates more Asian associated, for both proportion of occupations and for relative distance. $p < .05$ and with r-squared$ = .069$.} 
		\label{fig:appscatteroccasians}
	\end{subfigure}
\end{figure}

\begin{table}[H]
	\begin{center}
		\begin{tabular}{lclc}
			\toprule
			\textbf{Dep. Variable:}               & Hispanic Occupation Proportion & \textbf{  R-squared:         } &     0.277   \\
			\textbf{Model:}                       &              OLS               & \textbf{  Adj. R-squared:    } &     0.266   \\
			\textbf{Method:}                      &         Least Squares          & \textbf{  F-statistic:       } &     23.81   \\
			\textbf{Date:}                        &        Sat, 09 Sep 2017        & \textbf{  Prob (F-statistic):} &  7.77e-06   \\
			\textbf{Time:}                        &            19:58:08            & \textbf{  Log-Likelihood:    } &   -59.188   \\
			\textbf{No. Observations:}            &                 64             & \textbf{  AIC:               } &     122.4   \\
			\textbf{Df Residuals:}                &                 62             & \textbf{  BIC:               } &     126.7   \\
			\textbf{Df Model:}                    &                  1             & \textbf{                     } &             \\
			\bottomrule
		\end{tabular}
		\begin{tabular}{lccccc}
			& \textbf{coef} & \textbf{std err} & \textbf{t} & \textbf{P$>$$|$t$|$} & \textbf{[95.0\% Conf. Int.]}  \\
			\midrule
			\textbf{Relative Hispanic Similarity} &      17.0433  &        3.493     &     4.880  &         0.000        &        10.062    24.025       \\
			\textbf{const}                        &      -2.0415  &        0.091     &   -22.357  &         0.000        &        -2.224    -1.859       \\
			\bottomrule
		\end{tabular}
		\begin{tabular}{lclc}
			\textbf{Omnibus:}       &  2.268 & \textbf{  Durbin-Watson:     } &    2.273  \\
			\textbf{Prob(Omnibus):} &  0.322 & \textbf{  Jarque-Bera (JB):  } &    1.562  \\
			\textbf{Skew:}          &  0.152 & \textbf{  Prob(JB):          } &    0.458  \\
			\textbf{Kurtosis:}      &  2.297 & \textbf{  Cond. No.          } &     45.1  \\
			\bottomrule
		\end{tabular}
		\caption{Regression table corresponding to Figure~\ref{fig:appscatterocchispanics}}
	\end{center}
\end{table}

\begin{table}[H]
	\begin{center}
		\begin{tabular}{lclc}
			\toprule
			\textbf{Dep. Variable:}            & Asian Occupation Proportion & \textbf{  R-squared:         } &     0.069   \\
			\textbf{Model:}                    &             OLS             & \textbf{  Adj. R-squared:    } &     0.054   \\
			\textbf{Method:}                   &        Least Squares        & \textbf{  F-statistic:       } &     4.595   \\
			\textbf{Date:}                     &       Sat, 09 Sep 2017      & \textbf{  Prob (F-statistic):} &   0.0360    \\
			\textbf{Time:}                     &           19:58:06          & \textbf{  Log-Likelihood:    } &   -68.506   \\
			\textbf{No. Observations:}         &                64           & \textbf{  AIC:               } &     141.0   \\
			\textbf{Df Residuals:}             &                62           & \textbf{  BIC:               } &     145.3   \\
			\textbf{Df Model:}                 &                 1           & \textbf{                     } &             \\
			\bottomrule
		\end{tabular}
		\begin{tabular}{lccccc}
			& \textbf{coef} & \textbf{std err} & \textbf{t} & \textbf{P$>$$|$t$|$} & \textbf{[95.0\% Conf. Int.]}  \\
			\midrule
			\textbf{Relative Asian Similarity} &       6.9750  &        3.254     &     2.144  &         0.036        &         0.471    13.479       \\
			\textbf{const}                     &      -2.9873  &        0.212     &   -14.073  &         0.000        &        -3.412    -2.563       \\
			\bottomrule
		\end{tabular}
		\begin{tabular}{lclc}
			\textbf{Omnibus:}       &  0.012 & \textbf{  Durbin-Watson:     } &    1.911  \\
			\textbf{Prob(Omnibus):} &  0.994 & \textbf{  Jarque-Bera (JB):  } &    0.079  \\
			\textbf{Skew:}          & -0.025 & \textbf{  Prob(JB):          } &    0.961  \\
			\textbf{Kurtosis:}      &  2.835 & \textbf{  Cond. No.          } &     36.4  \\
			\bottomrule
		\end{tabular}
		\caption{Regression table corresponding to Figure~\ref{fig:appscatteroccasians}}
	\end{center}
\end{table}

\FloatBarrier
\subsection{Dynamic Analysis}
\label{sec:appracedynamic}

\begin{figure}[H]
	\centering
	\includegraphics[width=\linewidth]{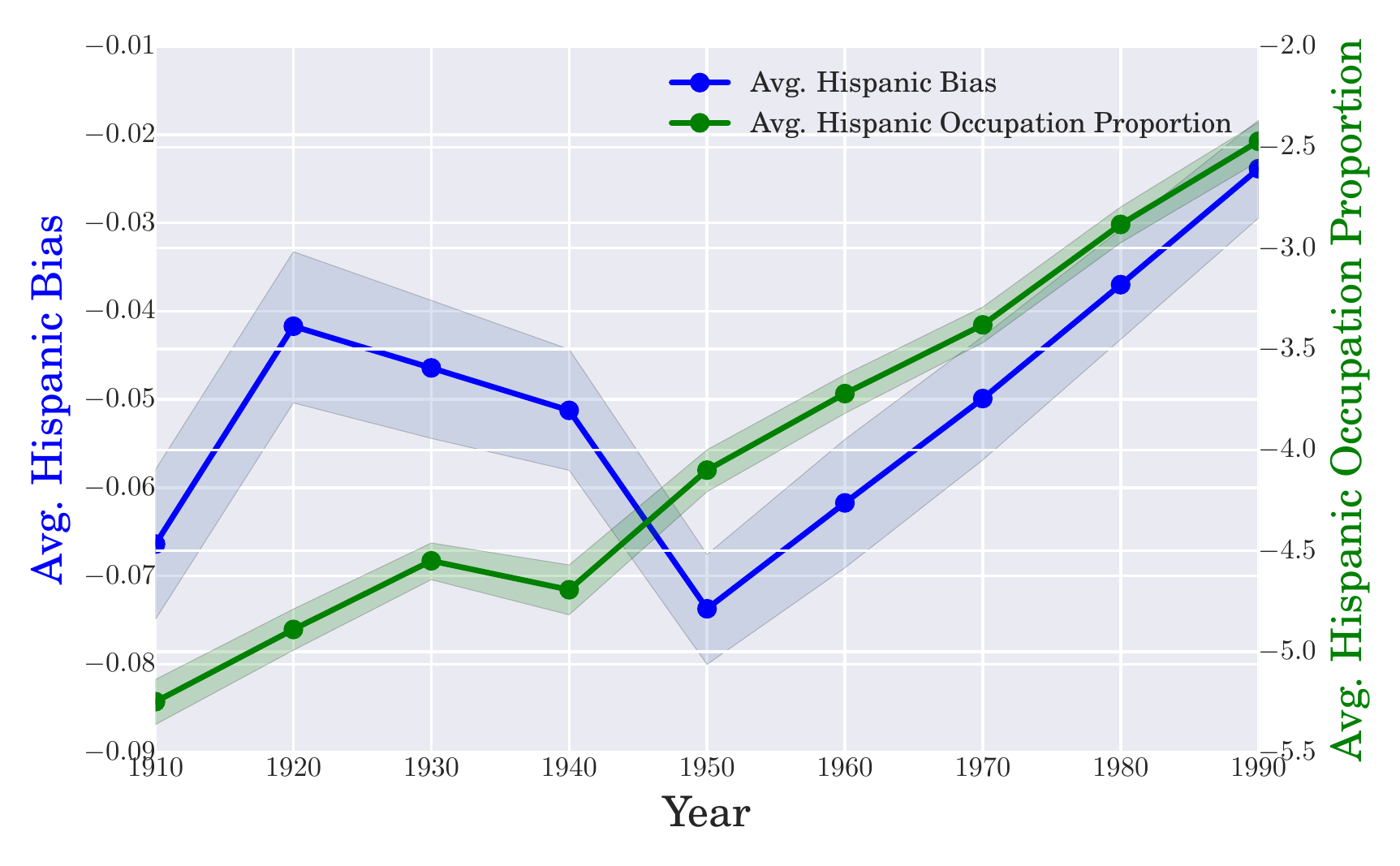}
	\caption{Ethnic (Hispanic vs White) bias over time in COHA dataset in occupations vs the average log proportion. In blue is the relative Hispanic bias in the SGNS embeddings, while in green is the average conditional log proportion of Hispanic in the occupation.}
	\label{fig:race_bias_over_time_appendix}
\end{figure}

\begin{figure}[H]
	\centering
	\includegraphics[width=.7\linewidth]{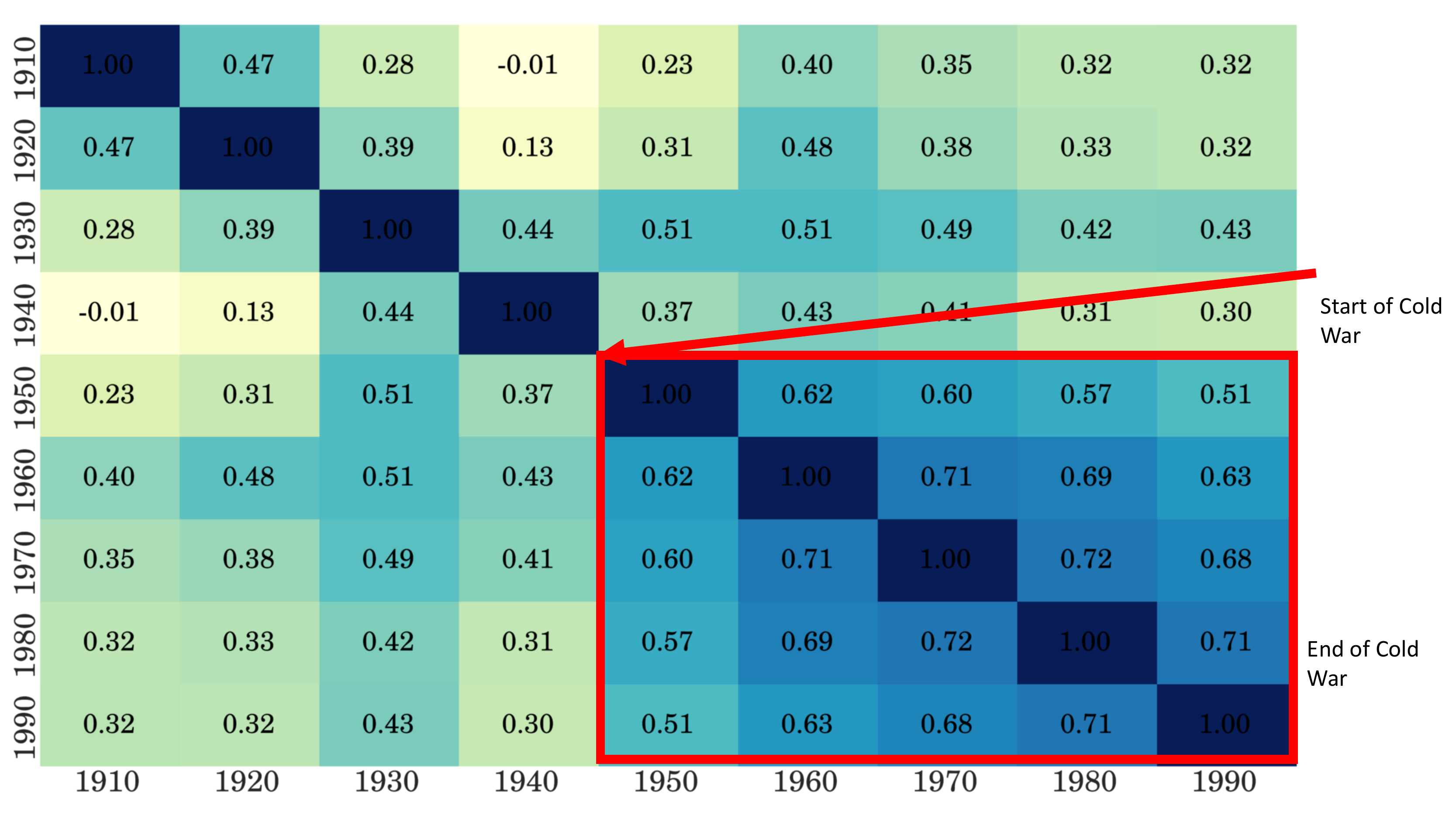}
	\caption{Pearson correlation in SGNS embedding Russian bias scores for adjectives over time between embeddings for each decade.}
	\label{fig:hispanicpearsoncorrelation}
\end{figure}
\begin{figure}[H]
	\centering
	\includegraphics[width=.7\linewidth]{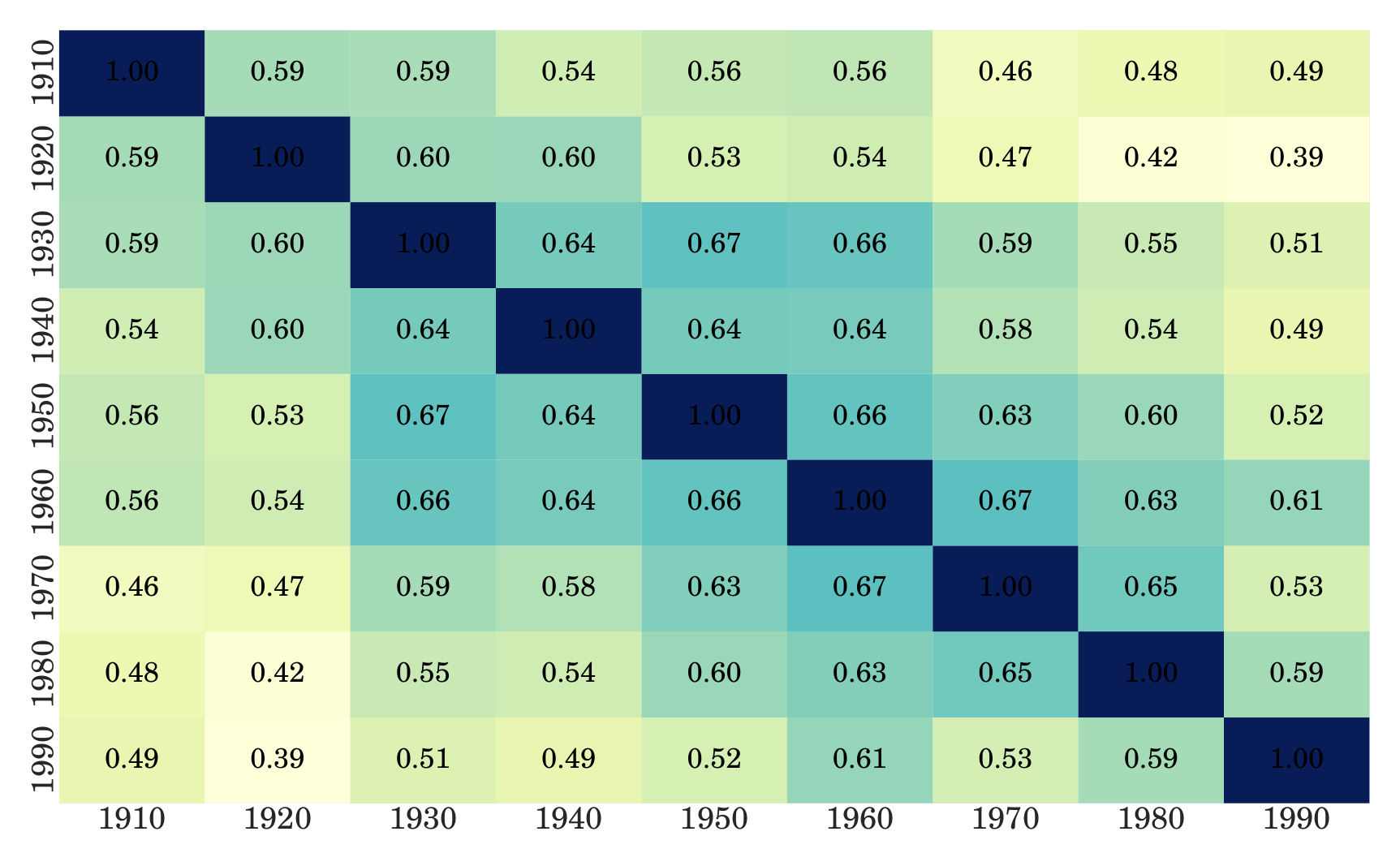}
	\caption{Pearson correlation in SGNS embedding Hispanic bias scores for adjectives over time between embeddings for each decade.}
	\label{fig:russianpearsoncorrelation}
\end{figure}

\begin{table}[H]
	\begin{center}
		\makebox[\textwidth][c]{
			\begin{tabular}{ccccccccc}
				1910 & 1920 & 1930 & 1940 & 1950 & 1960 & 1970 & 1980 & 1990 \\\hline
				irresponsible & mellow & hateful & solemn & disorganized & imprudent & cynical & superstitious & inhibited \\
				envious & relaxed & unchanging & reactive & outrageous & pedantic & solemn & upright & passive \\
				barbaric & haughty & oppressed & outrageous & pompous & irresponsible & mellow & providential & dissolute \\
				aggressive & tense & contemptible & bizarre & unstable & inoffensive & discontented & unstable & haughty \\
				transparent & hateful & steadfast & fanatical & effeminate & sensual & dogmatic & forceful & complacent \\
				monstrous & venomous & relaxed & assertive & unprincipled & venomous & aloof & appreciative & forceful \\
				hateful & stubborn & cruel & unprincipled & venomous & active & forgetful & dry & fixed \\
				cruel & pedantic & disorganized & barbaric & disobedient & inert & dominating & reactive & active \\
				greedy & transparent & brutal & haughty & predatory & callous & disconcerting & fixed & sensitive \\
				bizarre & compassionate & intolerant & disconcerting & boisterous & inhibited & inhibited & sensitive & hearty \\
		\end{tabular}}
		\caption{Top Asian (vs White) Adjectives over time by relative norm difference.}
		\label{tab:mostasianadjectives}
	\end{center}
\end{table}

\end{appendices}

\end{document}